\documentclass[letterpaper, 10 pt, conference]{format/ieeeconf}
\usepackage{amsmath}
\usepackage{amsfonts}
\usepackage{setspace}
\usepackage[space, compress, sort]{cite}

\usepackage{amsthm}
\usepackage{amssymb,stmaryrd}
\usepackage{mathtools}
\usepackage{tabularx}
\usepackage{graphicx}
\usepackage{subfigure}
\usepackage{enumerate}
\usepackage{float}
\usepackage{url}
\usepackage{verbatim}
\usepackage{cite}
\usepackage{booktabs} 
\usepackage{multirow}
\usepackage[linkcolor=black,citecolor=black,urlcolor=black,colorlinks=true]{hyperref}
\usepackage{bm}
\usepackage{empheq}
\IEEEoverridecommandlockouts
\usepackage{graphicx}
\usepackage{siunitx}
\usepackage{algorithm}
\usepackage{algpseudocode}
\usepackage{textcase}
\usepackage[tablename=TABLE]{caption}

\usepackage{tablefootnote}
\DeclareCaptionTextFormat{up}{\MakeTextUppercase{#1}}

\title{\LARGE \bf
MEGA-DAgger: Imitation Learning with Multiple Imperfect Experts
}

\author{Xiatao Sun$^{1*}$, Shuo Yang$^{2*}$, Mingyan Zhou$^{2}$, Kunpeng Liu$^{2}$, Rahul Mangharam$^{2}$% stops a space
\thanks{$^{1}$ Xiatao Sun is with the Department of Computer Science, Yale University, New Haven, CT 06510, USA (e-mail: xiatao.sun@yale.edu).}
\thanks{$^{2}$ Shuo Yang, Mingyan Zhou, Kunpeng Liu, Rahul Mangharam are with the Department of Electrical and Systems Engineering, University of Pennsylvania, Philadelphia, PA 19104, USA (e-mail: \{yangs1, derekzmy, kunpengl, rahulm\}@seas.upenn.edu).}
\thanks{$^*$Authors contributed equally.}}

\begin{document}
\maketitle

\begin{abstract}
Imitation learning has been widely applied to various autonomous systems thanks to recent development in interactive algorithms that address covariate shift and compounding errors induced by traditional approaches like behavior cloning.
However, existing interactive imitation learning methods assume access to one perfect expert.
Whereas in reality, it is more likely to have multiple imperfect experts instead.
In this paper, we propose MEGA-DAgger, a new DAgger variant that is suitable for interactive learning with multiple imperfect experts.
First, unsafe demonstrations are filtered while aggregating the training data, so the imperfect demonstrations have little influence when training the novice policy. Next, experts are evaluated and compared on scenarios-specific metrics to resolve the conflicted labels among experts.
Through experiments in autonomous racing scenarios, we demonstrate that policy learned using MEGA-DAgger can outperform both experts and policies learned using the state-of-the-art interactive imitation learning algorithms such as Human-Gated DAgger. The supplementary video can be found at \url{https://youtu.be/wPCht31MHrw}.
\end{abstract}

\IEEEpeerreviewmaketitle

\section{Introduction}

Learning-based control methods have shown successful applications in complex robotic systems~\cite{pierson2017deep}. Of wide applicability is imitation learning\cite{argall2009survey, fang2019survey, le2022survey, bojarski2016end}, which only requires expert demonstrations which are easy to collect at scale.
Among various imitation learning techniques, interactive methods such as DAgger~\cite{ross2011reduction} and Human-Gated DAgger (HG-DAgger)~\cite{kelly2019hg} are increasingly popular as they can address the covariate shift and compounding error induced by naive behaviour cloning~\cite{ross2010efficient}. 

Interactive imitation learning (IL)~\cite{ross2011reduction, zhang2016query, menda2019ensembledagger, kelly2019hg, hoque2021lazydagger, hoque2021thriftydagger} essentially involves expert feedback intermittently during novice policy training.
For example, DAgger trains the novice using a mixture labels from expert policy and novice policy.
In order to learn a more effective policy from human expert, HG-DAgger extends DAgger by introducing a human gated function to decide when expert should take over.
Robot-gated methods, such as SafeDAgger~\cite{zhang2016query}, allow the robot to actively query the human expert and request interventions only when necessary.
Nevertheless, interactive IL methods generally have two key assumptions: 
\begin{enumerate}
    \item expert demonstration is perfect; and
    \item all demonstrations are from a single expert.
\end{enumerate}
The first assumption seldom holds in reality, since human expert usually make mistakes. For example, in 2021, over 323 human-driven vehicle crashes occurred each day in the state of Pennsylvania~\cite{penndot}. 
Driver error accounted for over $85\%$ of all traffic crashes, which implies that human experts are sometimes unreliable as drivers.

\begin{figure}
    \centering
    \includegraphics[width=0.9\columnwidth]{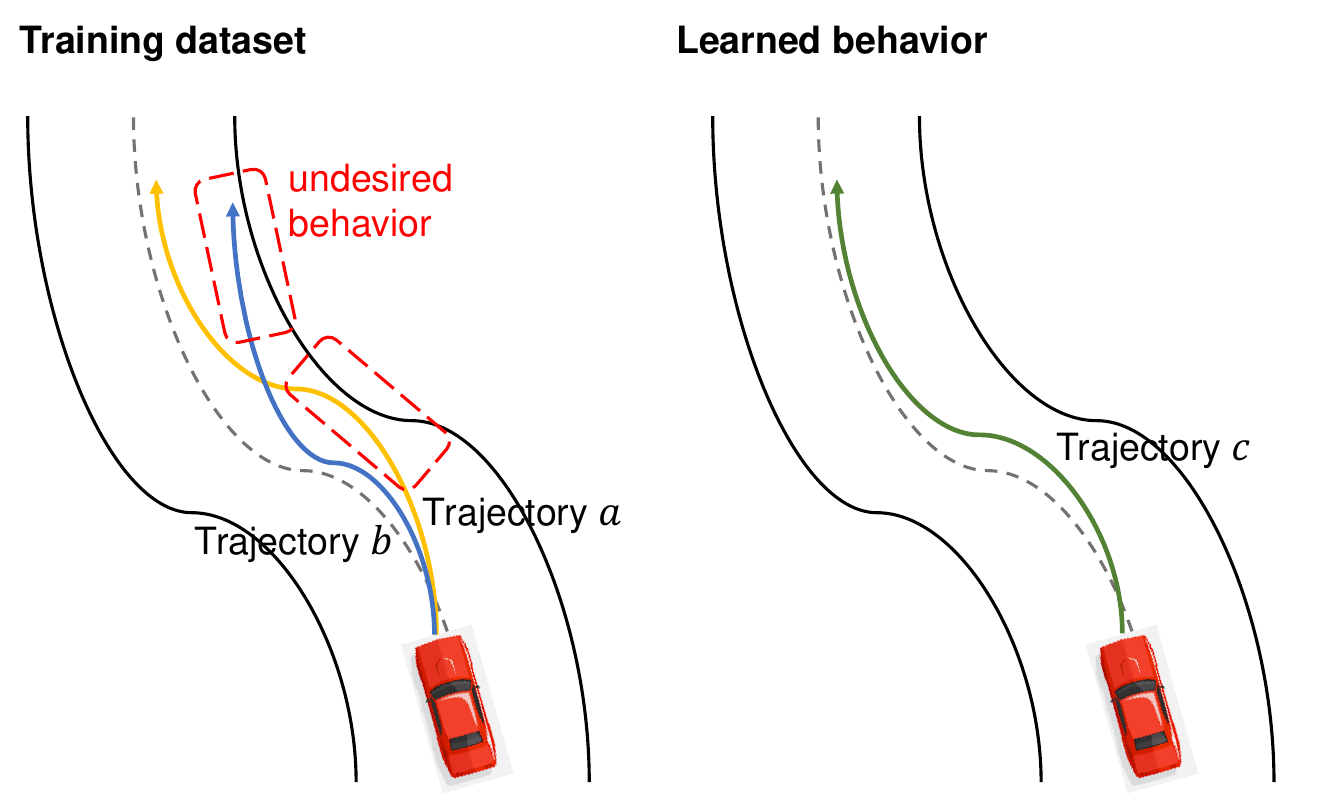}
    \caption{Illustration of learning from multiple imperfect experts. For example, two rollouts $a$ and $b$ (yellow and blue trajectories in the left figure) are from \emph{two different experts}, respectively. Each of them has undesired unsafe behavior (red box), and ideally we can learn complementary good behavior from both of them (green trajectory $c$ in the right figure).}
    \label{fig:concept}
    \vspace{-15pt}
\end{figure}

The second assumption is also invalid when there are multiple experts trying to teach a novice. For example, different drivers may have different driving policies. Some might have a performant but aggressive style, whereas others may choose to be safe but conservative. If one learns a policy from different experts simultaneously, demonstration labels provided by different experts may conflict. Thus, scenarios where multiple experts exist call for new techniques in imitation learning.
As illustrated in Figure~\ref{fig:concept}, two demonstrations from two experts are not perfect (left figure), and we hope to learn good behavior (right figure) from both of them and avoid undesired behavior.

% \vspace{-3cm}

In this work, we address the problem: \emph{How can we interactively learn from multiple imperfect experts}?
Towards this end, we propose Multiple-Expert-GAted DAgger (MEGA-DAgger), a DAgger variant that is designed for learning from multiple imperfect experts.
Specifically, we consider MEGA-DAgger for end-to-end autonomous racing, in which both safety and progress are crucial. 
We propose a control barrier function-based \emph{safety scorer} and filter unsafe expert demonstrations while aggregating new data.
At runtime, to resolve the multiple experts' action conflict, we evaluate each expert based on safety and progress scores and choose the best one. 
We show that our proposed solution outperforms existing DAgger variants and learns better-than-experts policy through experiments.
Note that, even if our framework has only been examined in the racing case, it can be easily applied to general autonomous systems with modified case-specific metrics.

\subsection{Related Work}
\textbf{Imitation learning for autonomous driving}: Previous work have successfully applied imitation learning to self-driving cars for safe driving, see, e.g., \cite{le2022survey}, \cite{abbeel2004apprenticeship, pan2017agile, zhou2021exploring}. Few studies adopt imitation learning policy for autonomous racing scenarios which need to maintain performance in addition to safety. \cite{sun2022benchmark} provides a benchmark for autonomous racing using imitation learning. However, these assume a single expert and cannot handle the multiple experts case. In addition, they also have the assumption that expert demonstrations are perfect, which is relaxed in this work.

\textbf{Imitation learning with imperfect expert}: learning from imperfect demonstrations is also studied in the past few years, see, e.g., \cite{wu2019imitation, brown2020better, wang2021learning, brown2019extrapolating}. These efforts focus on learning from imperfect demonstrations from single expert using inverse reinforcement learning. Instead, we consider interactive imitation learning with multiple imperfect experts for safety-critical autonomous systems in this paper. 

\textbf{Imitation learning with multiple experts}: when there are multiple experts, it is natural to raise a question: which expert should we select? This problem has been studied in the classification setting with the assumption that the labeled dataset is provided beforehand, see, e.g, \cite{raykar2009supervised, raykar2010learning}. However, in our work, the training process is interactive with unlabeled experts for complex autonomous systems, which makes existing literature inapplicable.

The contributions of this paper are summarized below:
\begin{enumerate}
    \item We propose Multiple-Expert-GAted DAgger (MEGA-DAgger), a new DAgger variant used for interactive imitation learning from multiple imperfect experts;
    \item We develop a data filter that can strategically truncate undesired demonstrations. This significantly improves the safety in autonomous racing scenarios compared with existing DAgger variants;
    \item We provide metrics to evaluate each expert and we empirically demonstrate that MEGA-DAgger can learn policies that outperform both experts and policies learned using HG-DAgger, the state-of-the-art interactive imitation learning algorithm.
\end{enumerate}

\section{Background}

\subsection{DAgger and HG-DAgger}
\emph{DAgger} is an interactive imitation learning algorithm that aggregates new data by running the expert policy and novice policy simultaneously ~\cite{ross2011reduction}. Specifically, at each integration $i$, new training data $\mathcal{D}_i$ is generated by:
\begin{align}
    \pi_i(x_t)=\phi_i \pi_{exp}(x_t) + (1-\phi_i) \pi_{N_{i-1}}(x_t),
\end{align}
where $x_t$ is the state at time step $t$, $\phi_i\in[0, 1]$ is a probability, $\pi_{exp}$ is the expert policy, and $\pi_{N_{i-1}}$ is the trained novice policy at iteration $i-1$.
Then, one can aggregate the dataset by $\mathcal{D}\leftarrow\mathcal{D}\bigcup\mathcal{D}_i$, and new policy $\pi_{N_{i}}$ is trained on $\mathcal{D}$.
By allowing the novice to affect the sampling state distribution, DAgger can mitigate the covariate mismatch and compounding error caused by behavior cloning~\cite{ross2010efficient}.

\emph{Human-Gated DAgger (HG-DAgger)} is a DAgger variant proposed to be a more suitable interactive algorithm for the task of learning from human expert~\cite{kelly2019hg}. It mainly differs by proposing the new rollout generation method:
\begin{align}\label{hgdagger-rollout-eq}
    \pi_i(x_t)=g(x_t) \pi_{exp}(x_t) + (1-g(x_t)) \pi_{N_{i-1}}(x_t),
\end{align}
where $g(x_t)$ is a gating function with value 0 if state $x_t$ is safe and value 1 if state $x_t$ is unsafe.
HG-DAgger assumes the expert is optimal and has privileged information regarding the safety of the current state.
Thus, the expert takes control only if unsafe behavior rolled by novice policy is observed.
Compared with DAgger, HG-DAgger can achieve better sample efficiency and improved training stability.

\begin{figure}
    \centering
    \includegraphics[width=0.93\columnwidth]{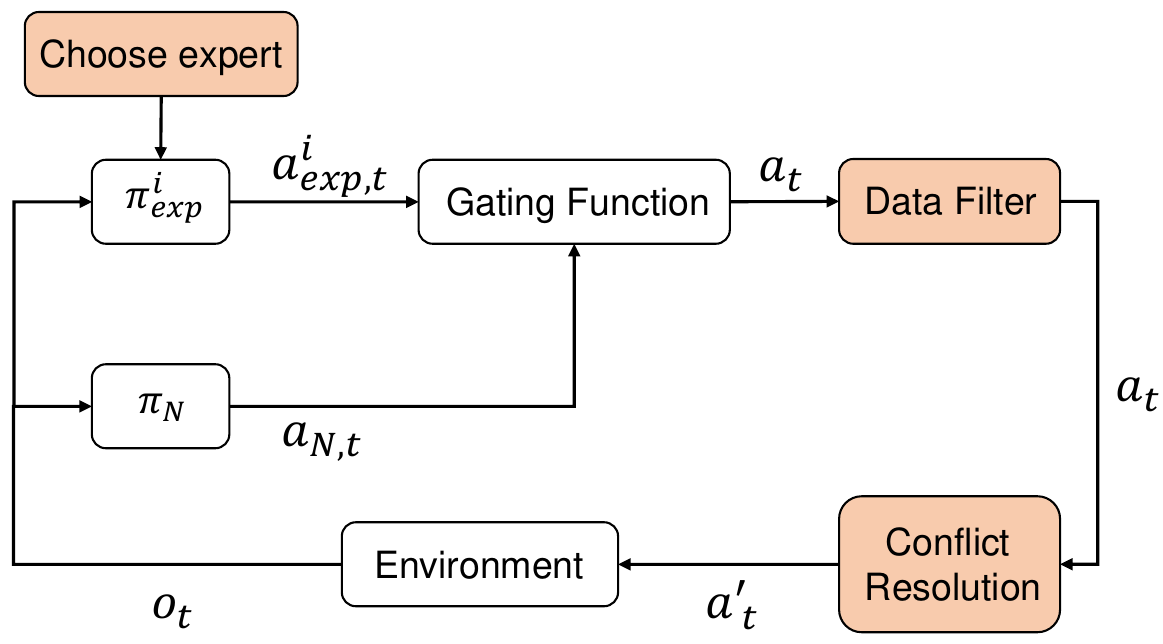}
    \caption{\small{Control loop for MEGA-DAgger. For each iteration, one expert should be chosen to be the dominant expert. Data Filter is used to remove unsafe demonstration and Conflict Resolution is used to eliminate actions conflict among experts.}}
    \label{fig:framework}
    \vspace{-15pt}
\end{figure}

Unlike DAgger and HG-DAgger, MEGA-DAgger is presented for multiple non-optimal experts case. 
As shown in Figure~\ref{fig:framework}, the overall loop is similar to HG-DAgger with three added blocks (shown in shaded color): in each iteration, first \emph{one expert is chosen} to be the dominant expert; 
then, \emph{data filter} removes unsafe demonstrations since experts are non-optimal and their behavior can also be unsafe;
finally, one needs to \emph{resolve  conflicts} from multiple experts.

\subsection{Evaluation using Autonomous Racing Benchmarks}
In this paper, we specifically study agent strategies for autonomous race cars in a head-to-head race~\cite{betz2022autonomous}. 
Autonomous racing provides a set of clearly specified metrics to balance safety and performance, making it a suitable scenario for research. 
Racing provides agents with the apparent goal of getting ahead of other agents and clear punishments if an agent crashes. 
Race cars of different scales are widely used for research and competition, such as full-scale Indy Autonomous Challenge~\cite{wischnewski2022indy}, Roborace~\cite{rieber2004roborace}, mid-scale Formula Student Driverless~\cite{zeilinger2017design}, and 1/10th-scale F1TENTH~\cite{o2020f1tenth}.

Learning-based control methods has attracted growing attention in autonomous racing, e.g., ~\cite{herman2021learn, kabzan2019learning}.
The imitation learning framework has been successfully applied~\cite{sun2022benchmark} since only human demonstrations are required in imitation learning and they are easy to collect.
In~\cite{sun2022benchmark}, only one expert is considered and it is assumed to be perfect.
However, racers have different and imperfect competition styles.
This motivates us to learn from multiple imperfect experts and finally have a better-than-experts policy.

% \subsection{Problem Formulation}
% Our goal is to learn a \emph{safe} and \emph{performant} policy (or better-than-expert policy?) from multiple weak experts' demonstrations for autonomous racing via imitation learning.

% We assume that we have access to multiple demonstrations sets $\mathcal{D}_1, \mathcal{D}_2, \cdots, \mathcal{D}_n$, which are generated by expert policies $\pi_1$, $\pi_2$, $\cdots, \pi_n$, respectively.
% A policy is a probability distribution over actions for given state. 

% We first define the lap time of a trajectory $\tau$ as:
% \begin{align}
%     \texttt{LapTime}(\tau)= \begin{cases} \text{time cost of } \tau, & \text{if } \tau \text{ safely finished the lap} \\
%                      \beta, & \text{if there is crash in } \tau
%        \end{cases},
% \end{align}
% where $\beta$ is a large number.
% Then, we define the \emph{cost} of a policy $\pi$ by $J(\pi) = \mathbb{E}_{\tau\sim\pi}{\texttt{LapTime}(\tau)}$.
\section{Methodology}

Algorithm \ref{alg:cap} provides an overview for the MEGA-DAgger algorithm. The algorithm starts with an empty global dataset $D$ and a randomly initialized novice policy $\pi_{N_0}$ from the class of all possible policies $\Pi$. Similar to HG-DAgger, MEGA-DAgger incorporates expert demonstrations $D_j$ from each rollout $j$ into $D$ incrementally at each iteration, with $j=\{1,...,M\}$, where $M$ is the number of experts. During a rollout, the novice performs inference and controls the ego vehicle until the expert notices that the novice enters an undesired region. The expert then intervenes and takes control under this circumstance, and provides action label $a$ for the current observation $o$. After guiding the ego vehicle back to the desired region, the control of the vehicle is handed over to the novice again. The pairs of observation and action for demonstrations are only recorded and collected when the expert intervenes and takes control.

% Oerview and compare with other DAgger variants
\begin{algorithm}
\caption{MEGA-DAgger}\label{alg:cap}
\begin{algorithmic}[1]
\Procedure {MEGA-DAgger}{$\pi^{1:M}_{exp}$}
\State Initialize $\mathcal{D}\gets \emptyset$
\State Initialize $\pi_{N_0}$ to any policy in $\Pi$
\For{\texttt{iteration} $i=1:K$}
    \For{\texttt{rollout} $j=1:M$ with expert $\pi^j_{exp}$}
        \For{\texttt{timestep} $t \in T$ of \texttt{rollout} $j$}
            \If{$\pi^j_{exp}$ takes control}
                \State $o\gets \texttt{rollout}_{i, j}^t$
                \State $a\gets \pi^j_{exp}(o)$%\Comment{This is a comment}
                \State $D_j\gets o, a$
                \State $D_j,\sigma_t \gets\Call{Data Filter}{D_j}$
                %\State $\sigma_{s}\gets\Call{data filter}{D_j, \alpha, \beta}$
                % alpha and beta are CBF threshold and number of removed steps respectively
                \EndIf
                % \If{\textit{Conflict}$((s, a), \mathcal{D})=$ False}
                %     \State Add $(s,a)$ to $\mathcal{D}$
                % \Else
                %     \State $(s', a')\gets$ conflicted data
                %     \If{Score$(s,a)>$Score$(s',a')$}
                %         \State Replace $a'$ with $a$
                %     \Else
                %         \State   Reinforce the weight of $(s,a)$ in $\mathcal{D}$?
                %     \EndIf
                % \EndIf
            
        \EndFor
        \State $D_j\gets\Call{Conflict Resolution}{D_j, D, \sigma_t}$
        \State $D\gets D\cup D_j$
    \EndFor
    \State Train $\pi_{N_{i}}$ on $\mathcal{D}$
\EndFor
\EndProcedure
\end{algorithmic}
\end{algorithm}

MEGA-DAgger considers learning from multiple imperfect experts $\pi^{1:M}_{exp}$, which is different from other DAgger variants that assume only having the access to a single optimal or near-optimal expert \cite{ross2011reduction, kelly2019hg}. In each iteration $i$, we let each expert $\pi^j_{exp}$ take turns to observe and to intervene if necessary in $\texttt{rollout}_{i, j}$.

\begin{figure}
    \centering
    \includegraphics[width=0.93\columnwidth]{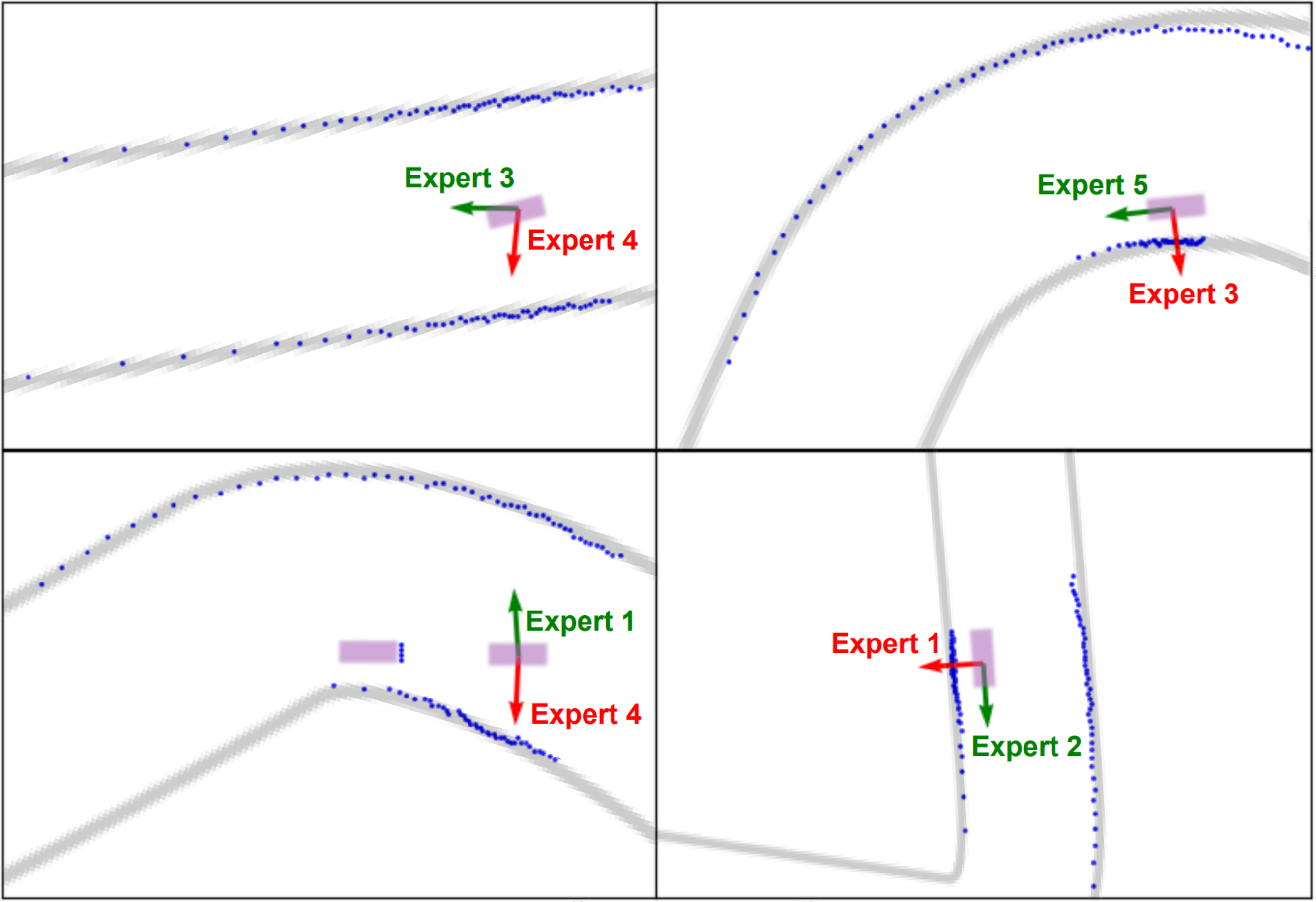}
    \caption{Illustration of conflicted labels from different experts. Blue dots represent hit points of LiDAR scans. Red and green arrows represent unit vectors of steering angles from labels. Pink rectangles represent ego and opponent vehicles. Grey lines represent the boundaries of the race tracks.}
    \label{fig:conflict_vis}
    \vspace{-10pt}
\end{figure}

\textbf{Challenges:} The scenario of multiple imperfect experts brings two major challenges. First, the expert demonstrations may not be safe. In HG-DAgger \cite{kelly2019hg}, safety is ensured by interventions from the optimal expert. Since such an optimal expert is missing in our context, unsafe demonstrations can be incorporated into the training dataset, which is detrimental to the novice policy and can potentially cause collisions when performing inference during autonomous racing.
Moreover, as shown in Figure~\ref{fig:conflict_vis}, different experts may provide drastically different labels for similar observations from adjacent states, which can consequently interfere with interpolations using the learned novice policy during inference.

% Unsafe data filter (CBF part)
\textbf{Solution for undesired demonstrations:} To mitigate the challenge of unsafe data, we design a data filter based on \emph{Control Barrier Function (CBF)}~\cite{ames2016control}. The data filter takes $D_j$ as input. It first checks the current LiDAR observation and gets the current ego position $(x_t^e, y_t^e)$, then outputs the CBF value $h(x_t^e, y_t^e)$, which is defined by 
\begin{align}
    h(x_t^e, y_t^e)=(x_t^{e} - x_t^{p})^2 - (y_t^{e} - y_t^{p})^2 - \alpha^2,
\end{align}
where $(x_t^{p}, y_t^{p})$ is the current obstacle (such as opponent agent, nearest boundary point) position and $\alpha$ is the corresponding minimal safe distance.
Leveraging the result of discrete-time CBF condition~\cite{zeng2021enhancing}, we define the safety score by:
\begin{align}
    \sigma_{t}=h(x_{t+1}^e, y_{t+1}^e)-(1-\gamma) h(x_{t}^e, y_{t}^e), 0<\gamma\le 1.
\end{align}
Note that higher $\sigma_{t}$ value indicates higher safety robustness\footnote{Constructing a valid CBF sometimes is expensive. In this work, we do not require valid CBF, as the CBF condition here is only used to provide a heuristic safety score.}. Therefore, the current rollout will be immediately terminated by the data filter once safety score $\sigma_t$ becomes negative. Since the ego vehicle has likely deviated from desired overtaking behavior several steps before it enters the unsafe region, $\beta$ number of previous steps are also truncated once the vehicle enters the unsafe region in order to remove as many undesired demonstrations as possible.

% Conflict resolution
\textbf{Solution for conflicted demonstrations:} To resolve the conflicted labels from different experts, a function for conflict resolution is executed after each rollout $j$ before the incorporation of $D_j$ into $D$. The conflict resolution takes $D_j$, $D$, and $\sigma_t$ as inputs. We use \emph{cosine similarity} to identify and select similar observations due to its wide applications in similarity detection for various sensors that are commonly seen in robotics, including LiDAR scans \cite{liu2019pedestrian, yang2023global} and RGBD camera \cite{zhou2015sharp, hu2022we}. To efficiently leverage parallel processing, the cosine similarities $\Theta$ between all observations $O_j$ in $D_j$ and all observations $O$ in $D$ is calculated by the dot product of $O$ and $O_j$ divided by element-wise multiplication of the Euclidean norms of $O$ and $O_j$:
$$
\Theta = \frac{O \cdot O_j}{\lVert O \rVert \odot \lVert O_j \rVert} .
$$
The indices of elements in $\Theta$ that are higher than a similarity threshold $\epsilon$ are selected to retrieve similar demonstrations from $D$ and $D_j$ for calculating the evaluation score $\omega_t$. 
In our autonomous racing context, $\omega_t$ for every similar demonstration can be calculated as the sum of the normalized safety score (safety indication) and normalized speed of the ego vehicle (progress indication)\footnote{One can also choose more complex score calculation methods such as assigning adapting weights to safety score and progress score depending on preference.}:
$$
\omega_t = \frac{\lVert  \sigma_t \rVert-\texttt{min}_t\lVert  \sigma_t \rVert}{\texttt{max}_t\lVert  \sigma_t \rVert-\texttt{min}_t\lVert  \sigma_t \rVert}  + \frac{\lVert  v_t \rVert-\texttt{min}_t\lVert  v_t \rVert}{\texttt{max}_t\lVert  v_t \rVert-\texttt{min}_t\lVert  v_t \rVert}.
$$
Within a group of similar demonstrations, the action label of the demonstration with the highest $\omega_t$ is then used to replace the action labels of all other similar demonstrations. After conflict resolution, $D_j$ is then merged with $D$. Finally, a policy $\pi_{N_{i+1}}$ is trained on $D$ at the end of iteration $i$.

\section{Experiments}

In this section, we provide experimental evaluations and demonstrate that our proposed MEGA-DAgger enjoys significant improved safety and performance compared with both experts and HG-DAgger. The experiment code, rosbag data, video links, etc, are available at \url{https://github.com/derekhanbaliq/f1tenth-MEGA-DAgger}.

\subsection{Experimental Setup}

To better understand the effect of each component in MEGA-DAgger on learning with multiple imperfect experts, we apply our method to learn overtaking behavior in a two-vehicle competitive autonomous racing scenario. We use the 2D racing simulator \texttt{f1tenth\_gym}\footnote{\url{https://github.com/f1tenth/f1tenth_gym}}\cite{o2020f1tenth} for our experiments.  Each vehicle in the simulator takes steering angle and speed as inputs and is equipped with a 2D planar LiDAR that outputs an array of laser scans with a length of 1080. Besides the LiDAR scan, the pose of each vehicle is also accessible at each step in the simulator. 
% During training, at each time step, the expert will take control if the difference in steering angle output by the novice and  the expert is greater than 0.1, or the difference in speed output by the novice and the expert is greater than 1. Otherwise, the vehicle is controlled by the novice.

For comparison, we choose to use HG-DAgger as our baseline since it is a state-of-the-art interactive imitation learning algorithm, and MEGA-DAgger is proposed as a step towards learning from imperfect experts based on it. During training and evaluation, each rollout is terminated either when the ego vehicle successfully overtakes the opponent, or the ego vehicle collides with the opponent or other obstacles. The environment is reset after the termination of a rollout. The \emph{percentage of overtakes} and \emph{percentage of collisions} are chosen as our main criteria for evaluation throughout our experiments. During the evaluation, each learned policy is tested for 100 rollouts. The number of rollouts that overtake or collide is recorded respectively to calculate the percentages by dividing the total number of rollouts.

We use the winning strategy lane switcher from the F1TENTH ICRA 2022 Race \cite{f1tenthicra2022results} both as the opponent and as the foundation for the experts planner. The lane switcher partitions the race track into two lanes. It keeps tracking one lane with Pure Pursuit until it encounters the opponent. Under this circumstance, it switches to the other unoccupied lane and tracks the lane using Pure Pursuit. Although the lane switcher is directly used as the opponent, for the expert planner, the lane switcher outputs reverse steering angles to generate undesired behaviors with a probability $P(U)$, where $U$ denotes the event of undesired behaviors. In this way, we are able to generate multiple different imperfect experts by setting different random seeds for $P(U)$.
The parameters in our experiments are listed in Table~\ref{table:parameter}.

\begin{table}
\centering
\small
\caption{\small{Values for hyperparameters. A two-layer MLP (multi-layer perceptron) with 256 hidden units is used as the novice policy.}}

\label{table:parameter}
\setlength{\tabcolsep}{26pt}
\begin{tabular}{*{2}{l}}
\toprule
\textbf{Hyperparameter} & \textbf{Value} \\
\midrule
Neural network structure & $2 \times 256$\\
Input dimension\tablefootnote{Instead of directly using the original LiDAR scan with length 1080 as input, the network takes a uniformly downsampled LiDAR array with a length of 108 as input to speed up training.} & 108\\
Evaluation rollouts number & 100\\
Minimal safe distance $\alpha$ & 0.42\\
Truncated step $\beta$\tablefootnote{More details of the choice of $\beta$ can be found in Appendix.} & 70\\
Cosine similarity threshold $\epsilon$ & 0.95\\
\bottomrule
\end{tabular}
\vspace{-10pt}
\end{table}

\subsection{Data Filter}

 To fairly evaluate the effect of the data filter, we first disable the conflict resolution function and only use HG-DAgger with our proposed data filter to learn from one expert with various $P(U)$ ranging from 0.1 to 1.0. In each trial, the novice policy learns from an expert with a fixed $P(U)$ for 1000 rollouts. For comparison, we also train novice policies in the same way with only HG-DAgger. To keep the amount of demonstrations the same for fair comparison, we randomly truncate the demonstrations when only using HG-DAgger to the same amount of demonstrations when using HG-DAgger with the proposed data filter after each rollout.
 The truncation ratios for different $P(U)$ are presented in Table~\ref{table:ratio}.

 \begin{figure}[h]
    \centering
    \includegraphics[width=1.0\columnwidth]{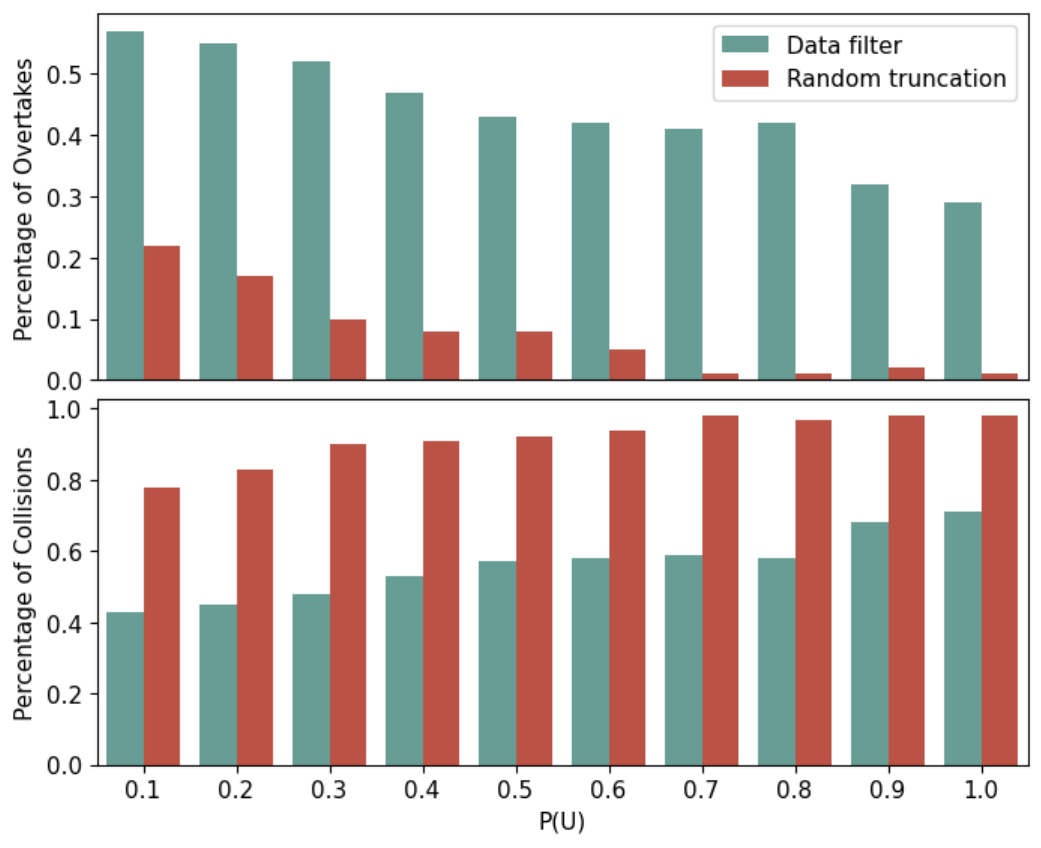}
    \caption{\small{The effect of the data filter on overtakes rate (above) and collisions rate (below), respectively. The results with different undesired behavior probability $P(U)$ are presented.}}
    \label{fig:diff_p_u}
\end{figure}

 As shown in Figure \ref{fig:diff_p_u},  using HG-DAgger with the proposed data filter shows significant improvement over using HG-DAgger with randomly truncated demonstrations. However, the performance of both the novice policies learned with each method gradually diminishes as $P(U)$ increases. Note that since $P(U)$ is only related to reversing the steering angle and does not guarantee collisions, the expert planner may still successfully overtake the opponent from another side of the opponent with the reversed steering angle and give good demonstrations. Therefore, even when $P(U)$ is equal to 1.0 and the data filter truncates bad demonstrations, the learned novice policy can still have around $30\%$ successful overtakes. This is also reflected the 18\% of the demonstrations that have passed through the data filter (as shown in Table~\ref{table:ratio}).

\begin{table}
\centering
\small
\caption{Ratios of removed demonstrations with different $P(U)$, where $r_\beta$ denotes the ratio of removed demonstrations from all collected demonstrations.}

\label{table:ratio}
\setlength{\tabcolsep}{11.5pt}
\begin{tabular}{*{6}{c}}
\toprule
$P(U)$ & $0.1$ & $0.2$ & $0.3$ & $0.4$ & $0.5$\\
\midrule
$r_{\beta}$ & $0.22$ & $0.25$ & $0.28$ & $0.32$ & $0.34$\\
\toprule
$P(U)$ & $0.6$ & $0.7$ & $0.8$ & $0.9$ & $1.0$\\
\midrule
$r_{\beta}$ & $0.38$ & $0.49$ & $0.57$ & $0.69$ & $0.82$\\
\bottomrule
\end{tabular}
\vspace{-10pt}
\end{table}

\subsection{Conflict Resolution}

% To evaluate the effect of the proposed conflict resolution method, we first generate 5 experts with different random seeds using the modified lane switcher with a fixed $P(U)$ of 0.5, so that the reversing steering angle is triggered at different steps to create undesired behaviors at different steps.
% Note that these random seeds are only applied to the experts and do not affect the MEGA-DAgger algorithm itself. In this way, the novice can learn from 5 imperfect experts that give different good and bad demonstrations at the same skill level using MEGA-DAgger.

To evaluate the effect of the proposed conflict resolution method, we generate 5 different experts based on the modified lane switcher with a fixed $P(U)$ value of 0.5.
For comparison, we train the novice policies using MEGA-DAgger, HG-DAgger with data filter, and HG-DAgger only on two different maps. When using HG-DAgger with and without the data filter, the novice only learns from one expert. Each method is used for training a randomly initialized novice policy during 1000 rollouts in total, with the network being saved and evaluated every 100 training rollouts for experiments. 5 trials are performed to calculate the 95\% confidence interval.

\begin{figure}[h]
    \centering
    \includegraphics[width=1.0\columnwidth]{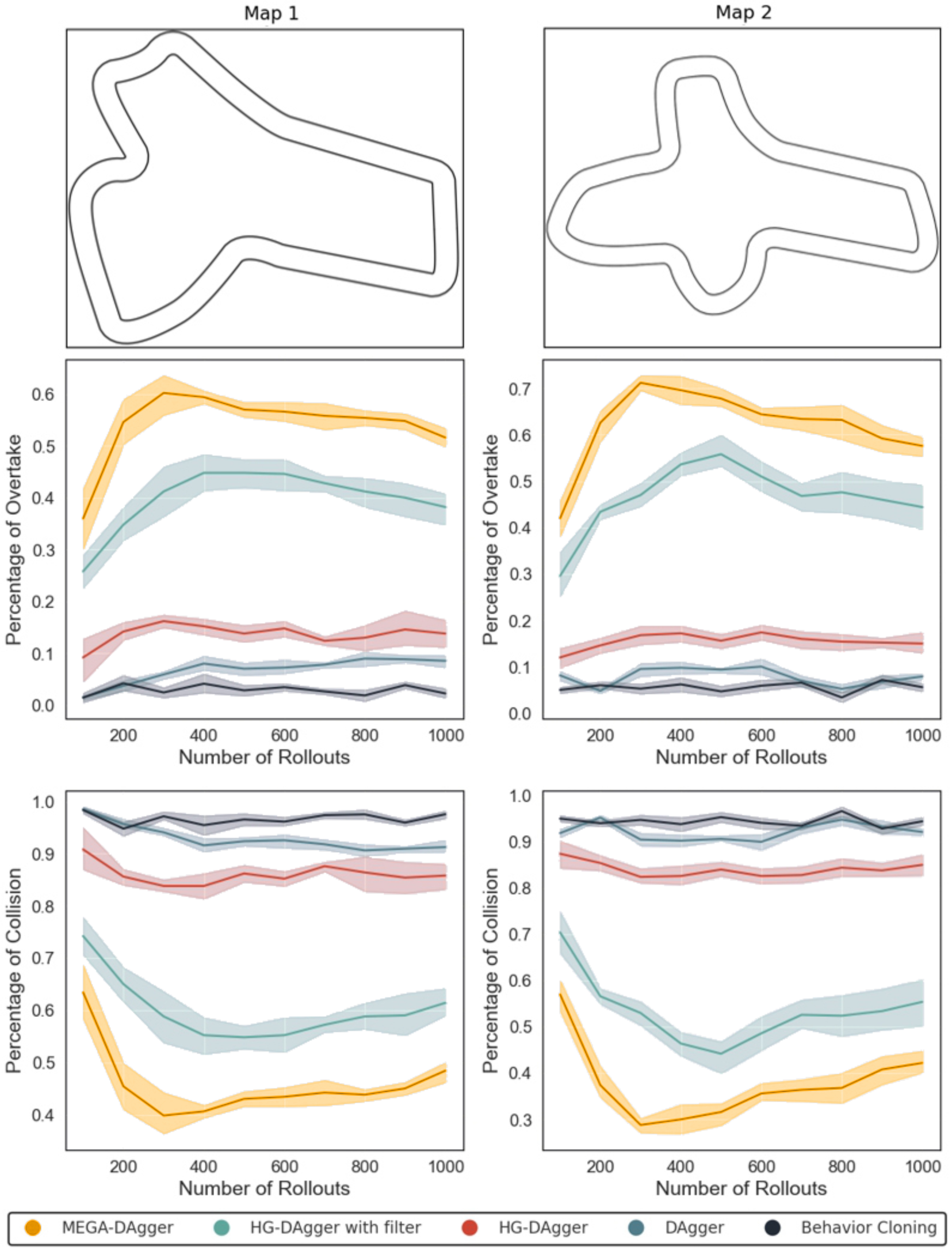}
    \caption{\small{Comparison of MEGA-DAgger, HG-DAgger with data filter, HG-DAgger, DAgger, and Behavior Cloning on two different maps. The left and right columns show the results on Map 1 and Map 2, respectively. Policy networks are saved every 100 training rollouts for evaluation.
    Each plot is an average of 5 experiments, and the shaded region represents 95\% confidence intervals. }}
    \label{fig:multi_single_filter_comp}
    \vspace{-10pt}
\end{figure}

\textbf{Better than HG-DAgger}: Figure \ref{fig:multi_single_filter_comp} shows the maps and the experimental results of learned policies. DAgger and behavior cloning perform worst regarding both overtaking and collision avoidance metrics. MEGA-DAgger has about 45\% average improvement on both overtaking and collision avoidance compared with vanilla HG-DAgger, and has about 15\% average improvement compared with HG-DAgger with data filter. When only using HG-DAgger, the novice can barely learn from demonstrations from multiple imperfect experts. Although compared with vanilla HG-DAgger, the data filter and conflict resolution functions demonstrate noticeable improvements on overtaking and collision avoidance, both metrics gradually decrease after reaching the peak at around 300 training rollouts. This indicates that MEGA-DAgger is able to reduce the amount of unsafe and conflict demonstrations rather than completely eliminating them. As more demonstrations are incorporated into the global dataset, the effect of MEGA-DAgger slowly decays, the cause of which we conjecture is that more undesired data is included during training.

\begin{figure}
    \centering
    \includegraphics[width=1.0\columnwidth]{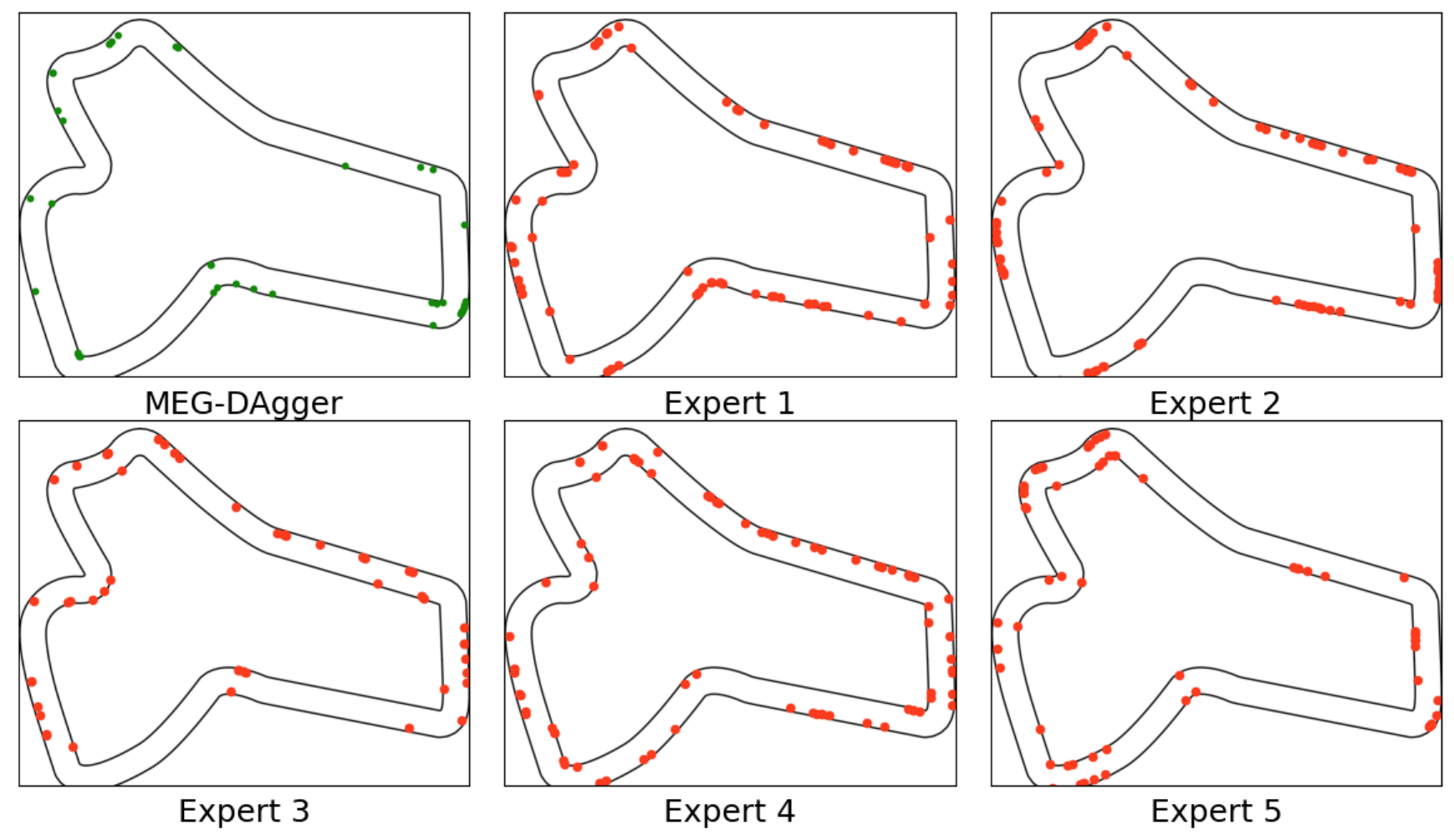}
    \caption{\small{Visualization of collision points caused by MEGA-DAgger policy and different experts over 200 evaluation rollouts on Map 1. The counts of collisions for MEGA-DAgger policy and Expert 1 to 5 are 40, 71, 79, 51, 79 and 63, respectively.}}
    \label{fig:collision_vis}
    \vspace{-10pt}
\end{figure}

\begin{table*}[t]
\small
\caption{Percentage of collisions and overtakes for the MEGA-DAgger policy and each expert. The means and standard deviations are calculated from 5 trials.}
\label{table:exp_nov_performance}
\setlength{\tabcolsep}{2.0pt}
\begin{tabular*}{\linewidth}{@{\extracolsep{\fill}} cccccccc }
\toprule
Metrics & MEGA-DAgger & Expert 1 & Expert 2 & Expert 3 & Expert 4 & Expert 5 & Experts Cumulative\\
\midrule
Collisions Percentage & $\bold{0.212 \pm 0.019}$ & $0.340 \pm 0.025$ & $0.401 \pm 0.033$ & $0.291 \pm 0.025$ & $0.392 \pm 0.028$ & $0.317 \pm 0.032$ & $0.348 \pm 0.051$ \\
Overtakes Percentage & $\bold{0.781 \pm 0.016}$ & $0.657 \pm 0.027$ & $0.594 \pm 0.036$ & $0.706 \pm 0.022$ & $0.605 \pm 0.024$ & $0.681 \pm 0.030$ & $0.649 \pm 0.051$\\
\bottomrule
\end{tabular*}
\end{table*}

\textbf{Better than experts}: We empirically attribute the improved performance of MEGA-DAgger over HG-DAgger with data filter to learning from complementary good demonstrations from different experts. 
By visualizing the collision points of a learned policy using MEGA-DAgger and all experts over 200 evaluation rollouts (as shown in Figure \ref{fig:collision_vis}), we can see that each expert frequently collides in different regions of the map, and the learned policy can learn complementary good behavior from them and have less collision.
Table~\ref{table:exp_nov_performance} shows that MEGA-DAgger is better than all experts and it has 13.6\% and 13.2\% average improvements on collision avoidance and overtaking, respectively.
Also, we find that the trained policy from MEGA-DAgger is more stable (smaller standard deviations) than experts. 
Since MEGA-DAgger is able to resolve conflicts by picking the best action under similar observations as illustrated in Figure \ref{fig:conflict_vis}, it is able to effectively leverage the complementary nature of the collision points, resolve conflicts, and learn a better-than-experts policy as a result.

\subsection{Effect of Similarity Threshold $\epsilon$}

To investigate the impact of the cosine similarity threshold $\epsilon$ in MEGA-DAgger, we conduct experiments by training policies with varying $\epsilon$ values and visualize the interplay between $\epsilon$, the number of training rollouts, and performance metrics (percentage of collisions and overtakes). The results, presented in Figures \ref{fig:cosine_sim_collision} and \ref{fig:cosine_sim_overtake} in Appendix, reveal that the choice of $\epsilon$ significantly influences the policy's performance. In our scenario, setting $\epsilon$ to 0.95 yields the best results. When $\epsilon$ is set to 1, the conflict resolution mechanism is disabled. Conversely, excessively decreasing $\epsilon$ leads to the policy failing to learn the desired behavior, as the conflict resolution incorrectly identifies dissimilar states as similar.

These findings can be interpreted in light of our observations from Figure \ref{fig:collision_vis}, which illustrates the complementary nature of collision points among different experts. We hypothesize that a carefully chosen $\epsilon$ value allows MEGA-DAgger to leverage the strengths of various experts by selectively resolving conflicts, resulting in a more robust policy. However, an overly aggressive $\epsilon$ value may cause MEGA-DAgger to overestimate the similarity between states, potentially leading to a suboptimal policy. Therefore, the choice of $\epsilon$ represents a trade-off between leveraging the diversity of expert demonstrations and maintaining the necessary level of state discrimination for effective learning.

\subsection{Real-World Experiments}

\begin{figure*}[t]
    \centering
    \subfigure[]{\includegraphics[width=0.325\textwidth]{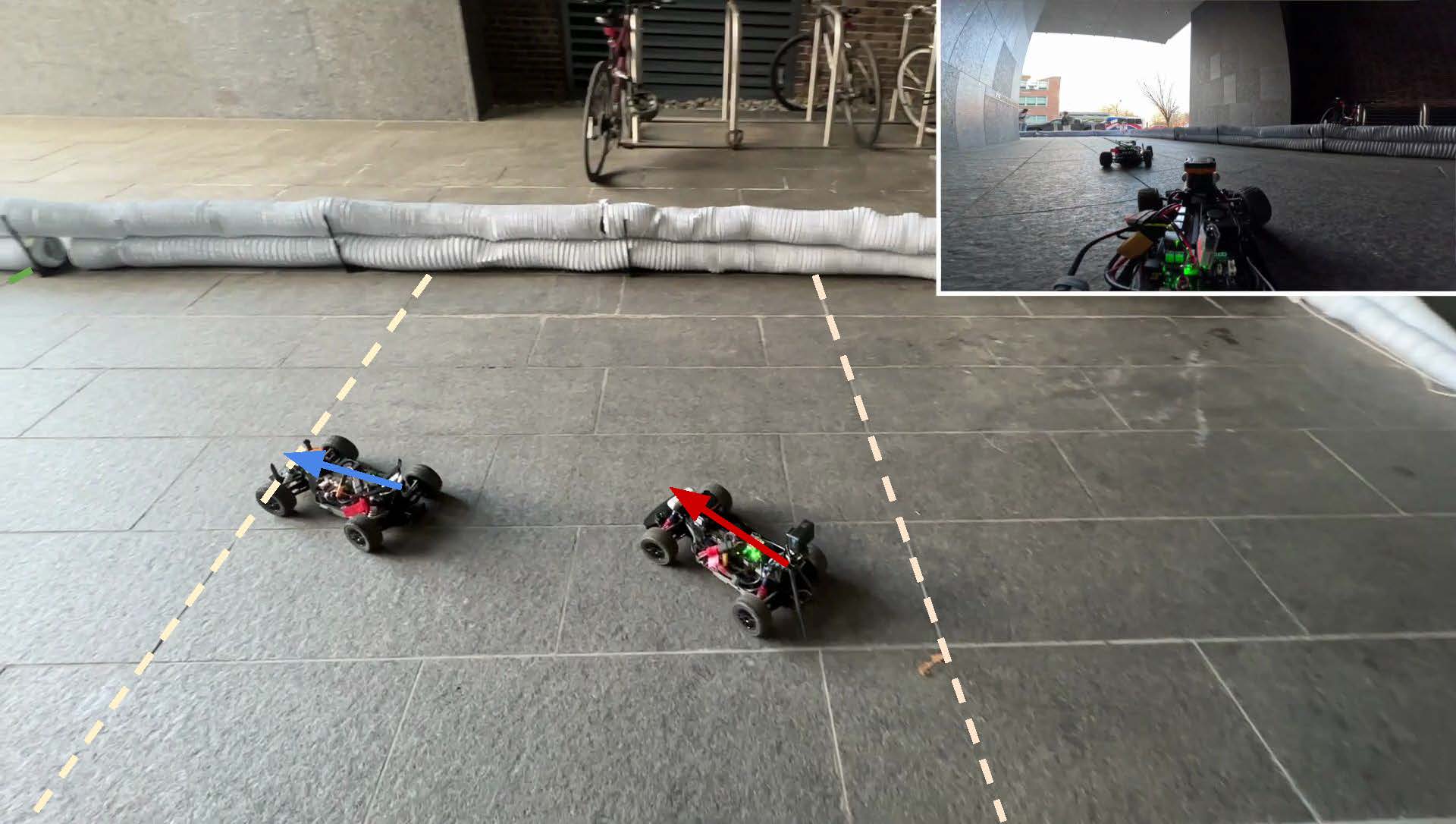}} 
    \subfigure[]{\includegraphics[width=0.325\textwidth]{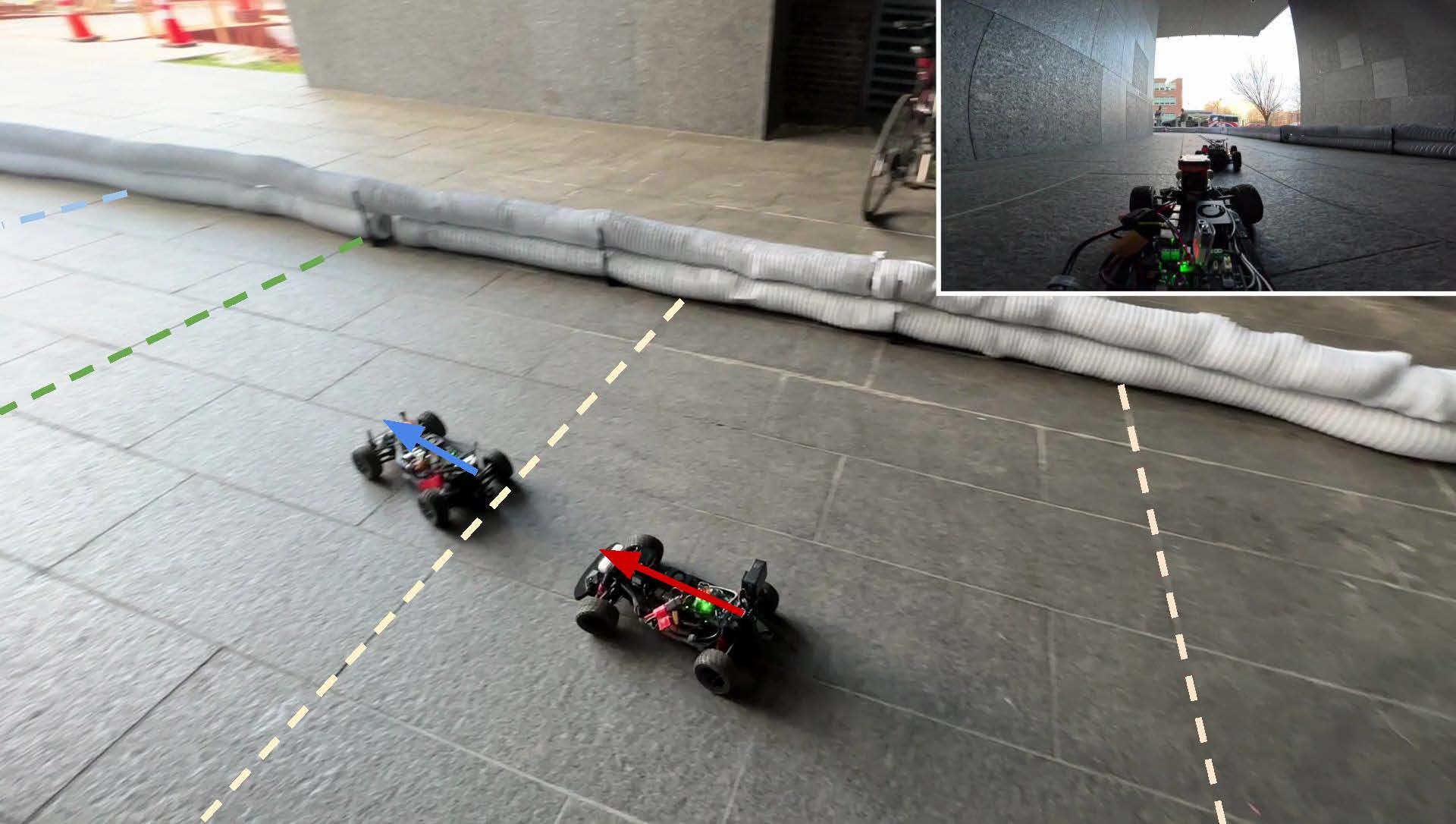}} 
    \subfigure[]{\includegraphics[width=0.325\textwidth]{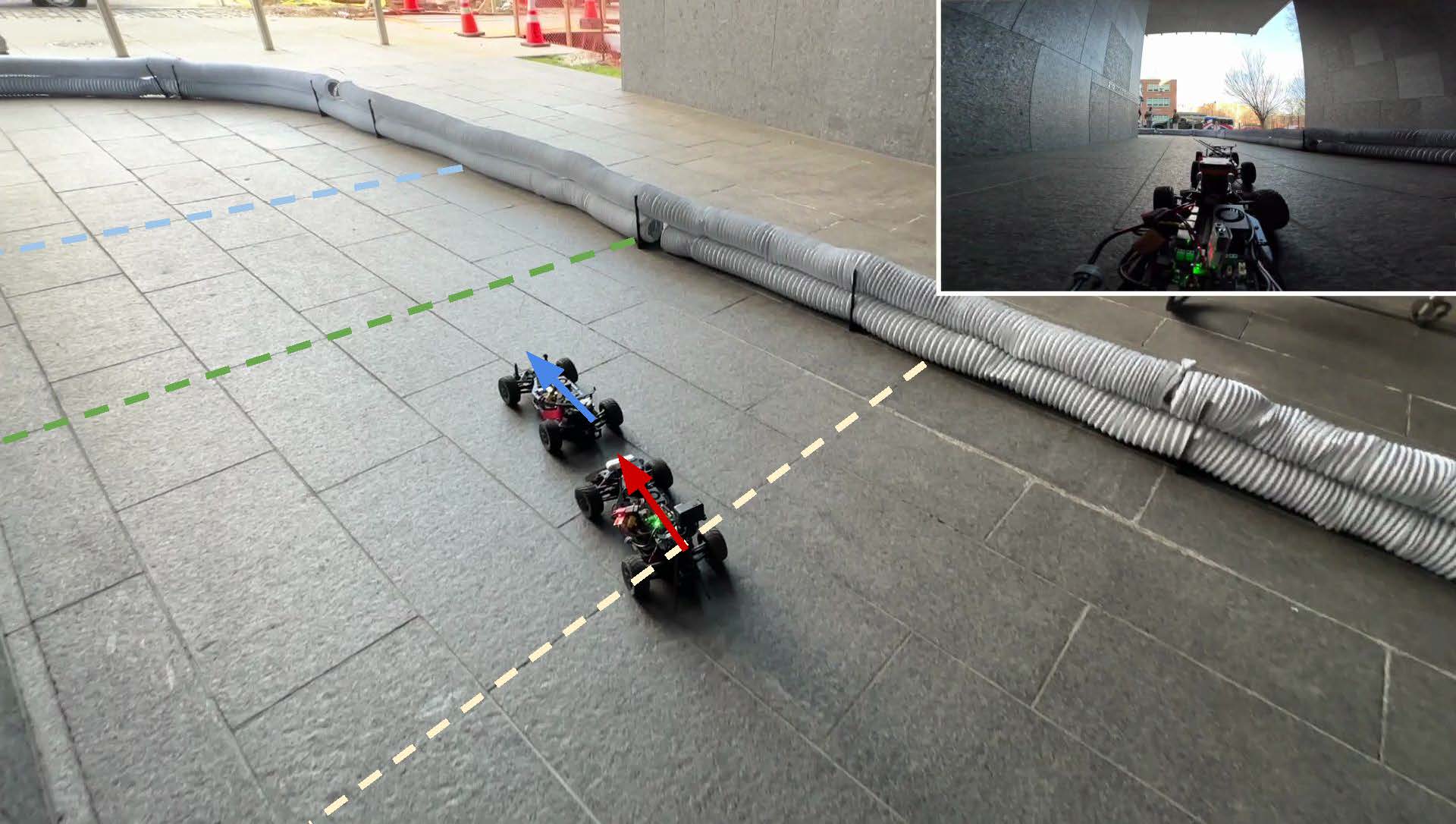}} 
    \subfigure[]{\includegraphics[width=0.325\textwidth]{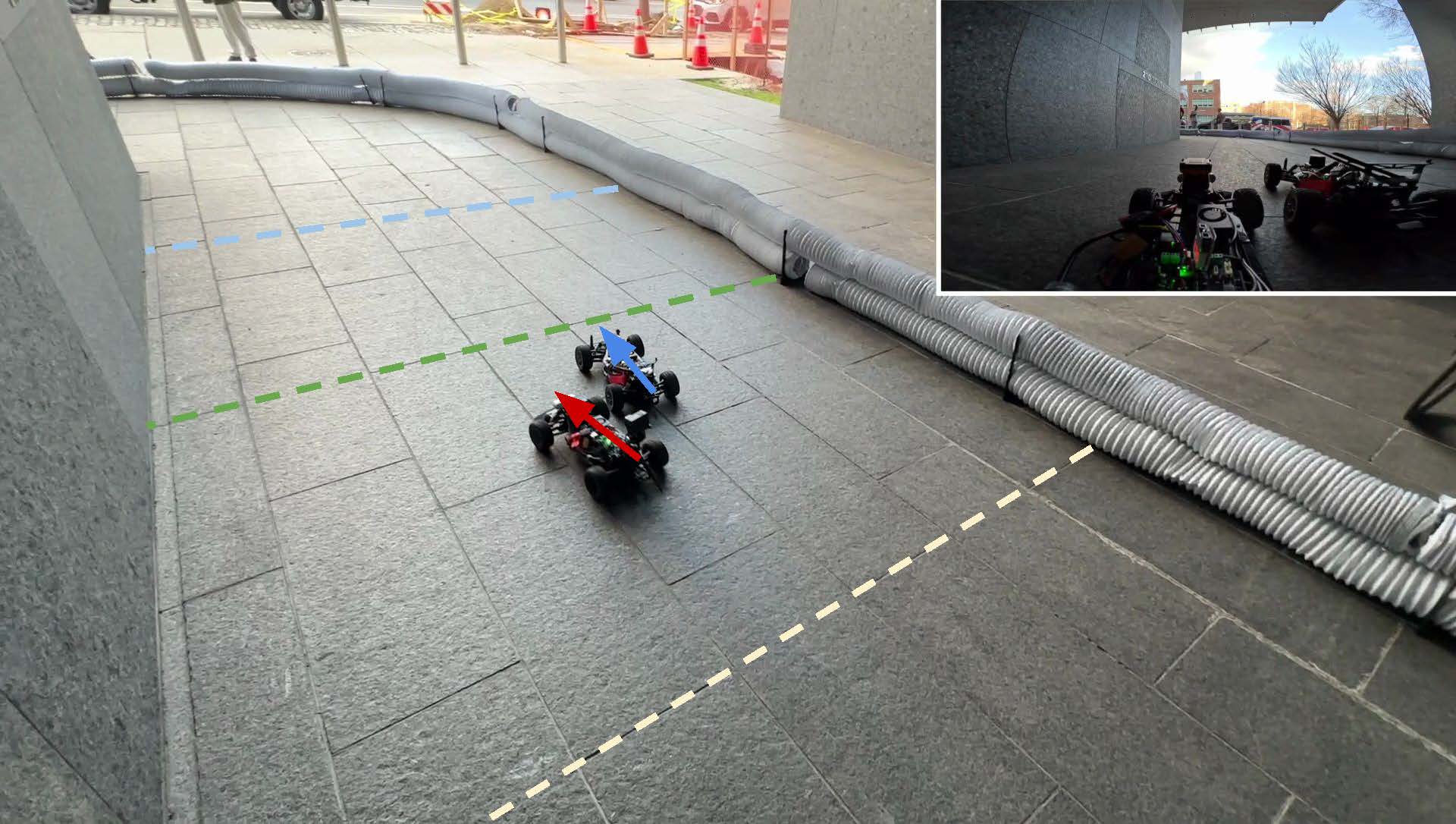}}
    \subfigure[]{\includegraphics[width=0.325\textwidth]{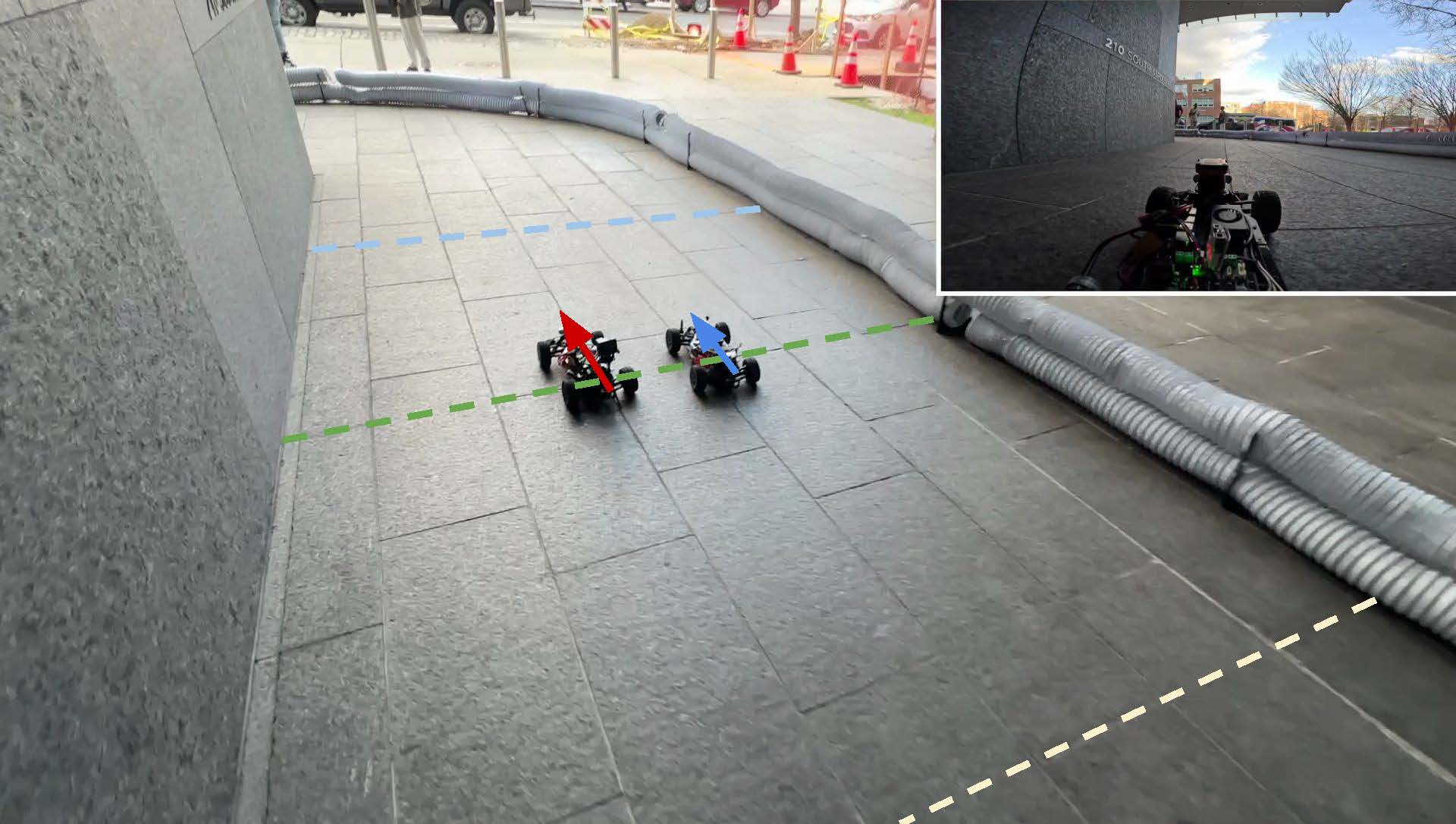}}
    \subfigure[]{\includegraphics[width=0.325\textwidth]{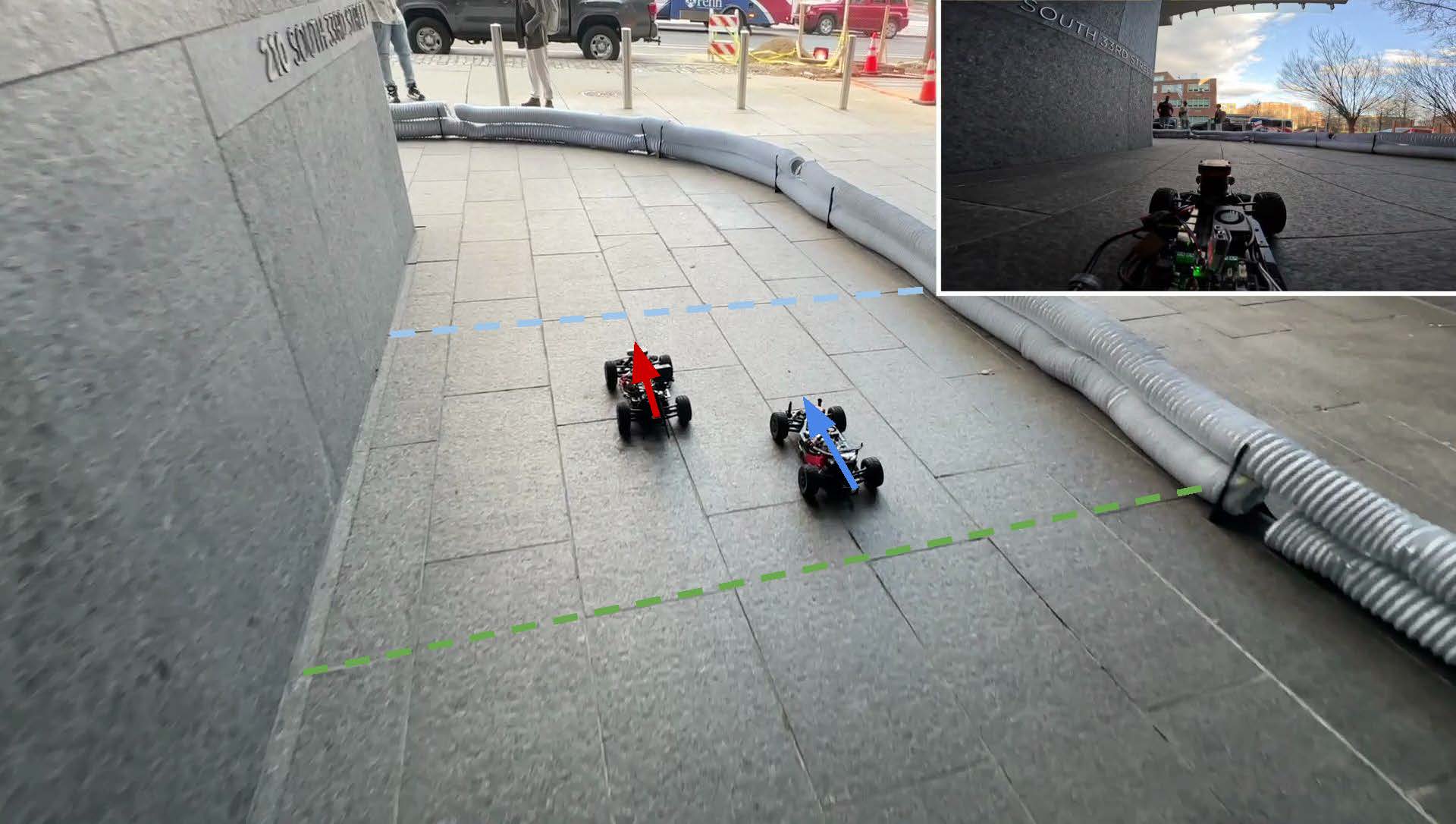}}
    \vspace{-10pt}
    \caption{\small{Consecutive frames with equal time intervals from (a) to (f) depict the safe overtaking process of MEGA-DAgger on the F1TENTH platform along a straight lane. The red and blue arrows indicate the estimated poses of the ego car and opponent car respectively. Equidistant dash lines represent the longitudinal progress of the track. The picture-in-picture in each frame shows the ego car's perspective. }}
    \vspace{-10pt}
    \label{fig:f110_real_car_exp}
\end{figure*}

\begin{figure}[h]
    \centering
    \subfigure[Sim]{
        \includegraphics[width=0.45\columnwidth]{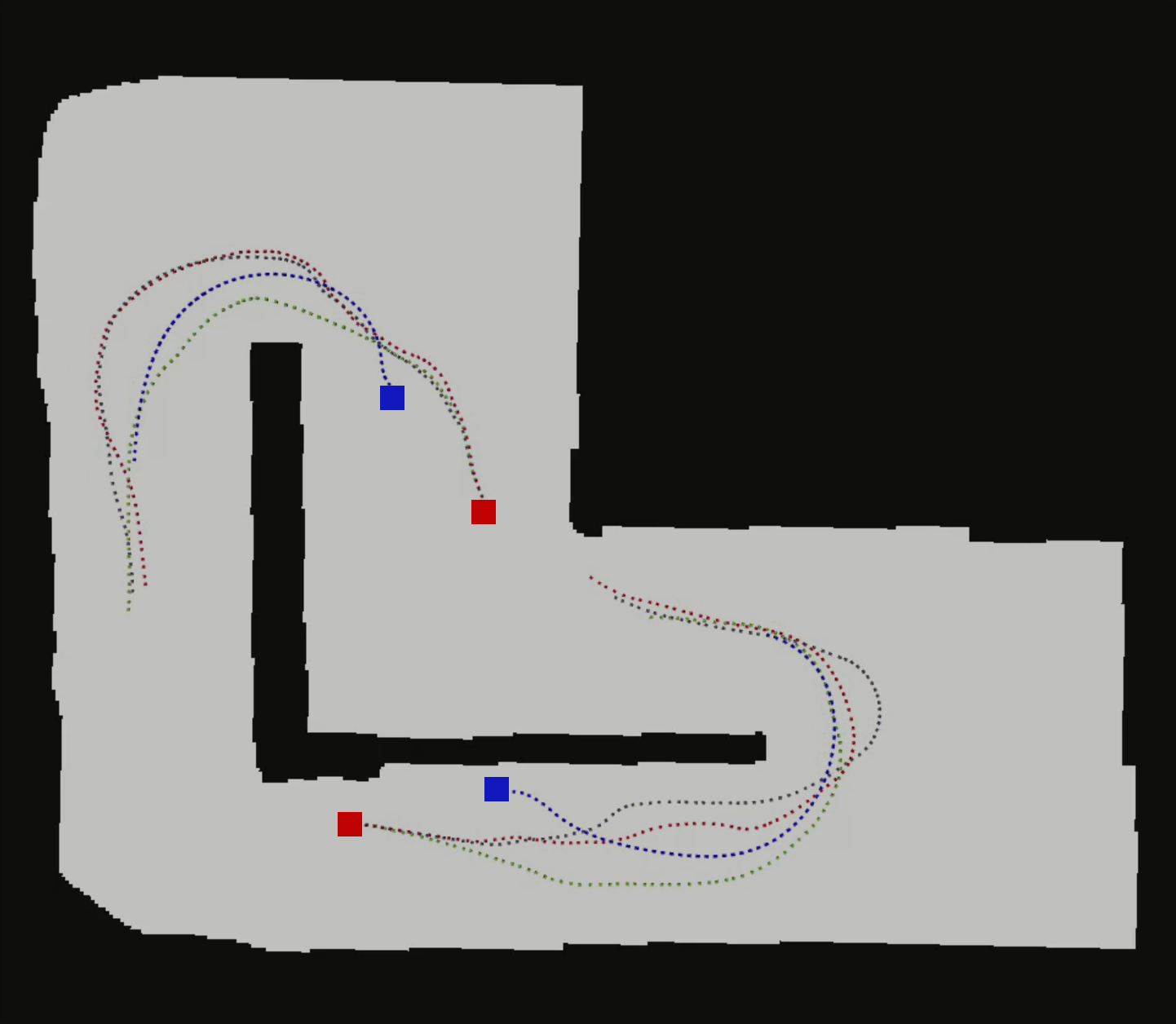}
        \label{fig:sim}
    } 
    \subfigure[Real]{
        \includegraphics[width=0.45\columnwidth]{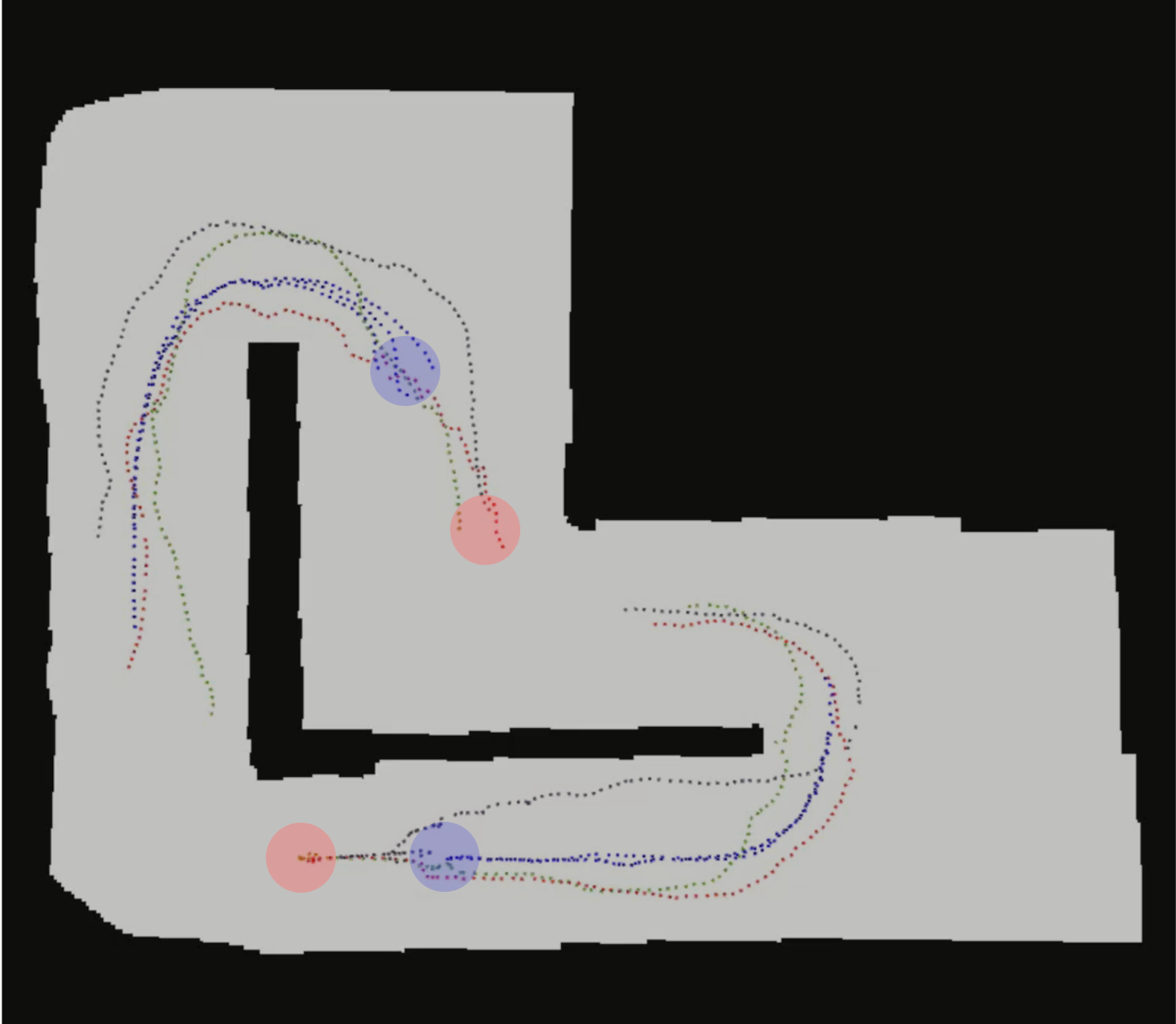}
        \label{fig:real}
    } 
    \caption{\small{Overtaking trajectories of the ego car (red, green, purple) using MEGA-DAgger and trajectories of the opponent car (blue) in the F1TENTH simulation and the real-world scenarios in RViz2. Red and blue marks show the initial positions of the ego car and the opponent car separately.} }
    \label{fig:f110_sim2real}
\end{figure}

% ego car & oppo car
We implement the MEGA-DAgger on the F1TENTH autonomous racing platform. The algorithm is deployed on the ego vehicle, performing cruising and overtaking safely. The opponent vehicle employs pure pursuit with a lower speed for path-tracking purposes. 
% sim2real - overtaking
We evaluate the MEGA-DAgger performances in both simulation and real-world scenarios under the ROS 2 environment. For both scenarios, the traces of both the ego car and the opponent car are recorded. As illustrated in Figure \ref{fig:f110_sim2real}, MEGA-DAgger can perform reliable strategies even under the sim-to-real gap. 
% real - overtaking
As shown in Figure \ref{fig:f110_real_car_exp}, the ego vehicle successfully completes the driving, and it demonstrates safe overtaking maneuvers by effectively passing the opponent.

\section{Conclusion and Discussion}
While interactive imitation learning methods, such as DAgger and its variants, have been successfully applied to many autonomous systems, they all assume the access to one optimal expert. 
However, it is more likely to only have access to multiple non-optimal experts. 
In this paper, we study how to make effective use of these experts through interactive imitation learning.
Specifically, MEGA-DAgger, a new DAgger variant, is proposed to filter unsafe demonstrations and resolve experts conflict.
Through thorough experiments on end-to-end autonomous racing, we demonstrate that MEGA-DAgger has improved safety and performance relative to HG-DAgger.
We also show that MEGA-DAgger can learn a better-than-experts policy.

It is worth noting that we use both the progress score and safety score to heuristically evaluate demonstrations, but they are not used as training feedback.
This is different from \emph{reward function} in reinforcement learning context, which is used to guide the training process and usually needs to be carefully designed.
One interesting direction for future work is to automatically learn confidence scores to evaluate and compare actions from experts.
Also, we plan to conduct experiments on real-world autonomous vehicles and trying to reduce the sim-to-real gap.

\section*{Appendix}\label{sec:appendix}
\begin{figure}[h]
    \centering
    \includegraphics[width=1.0\columnwidth]{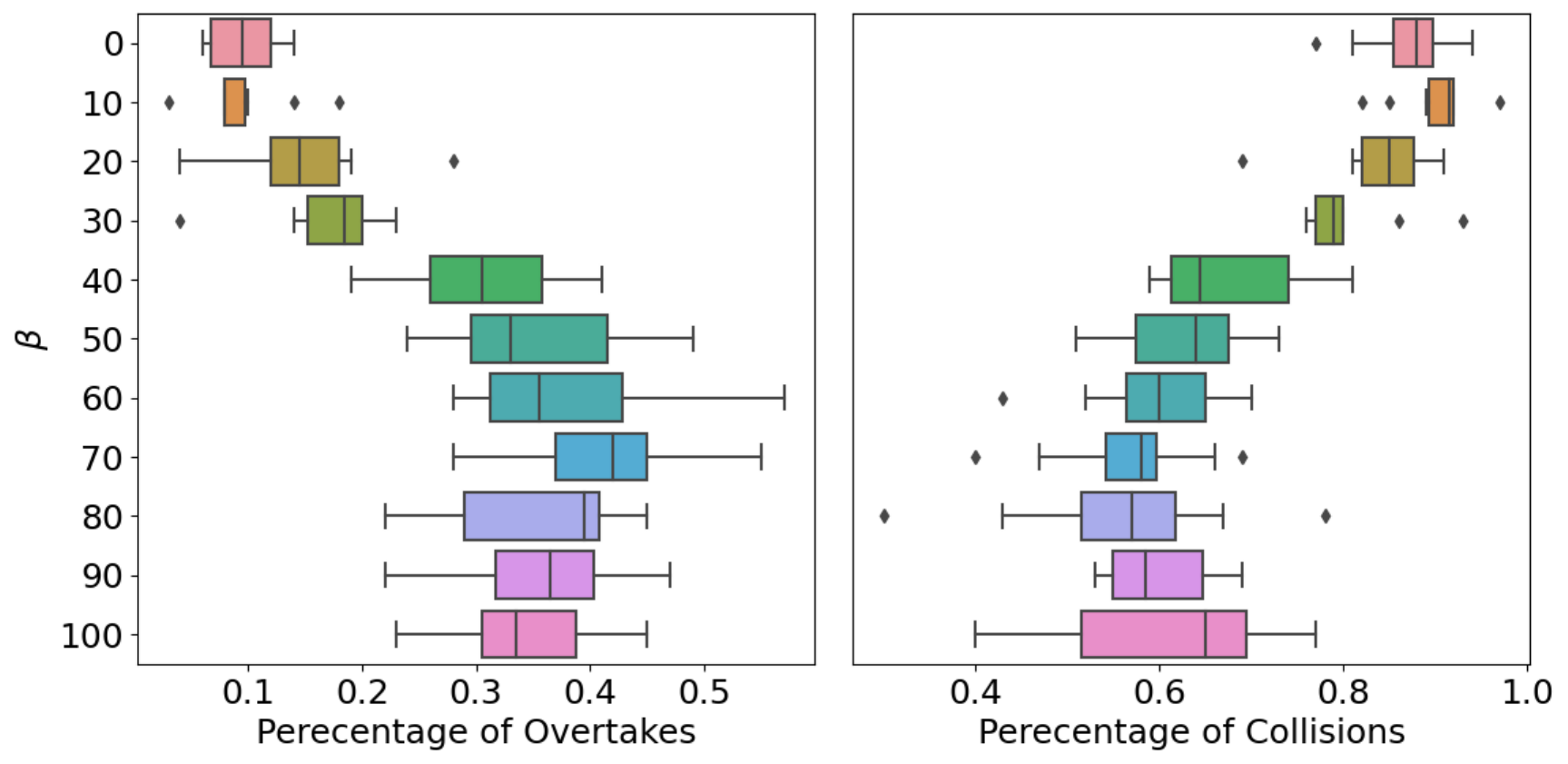}
    \caption{\small{Evaluation for preliminary trained policies to find the most appropriate $\beta$ (truncation steps for unsafe demonstration) for the overtaking task in autonomous racing. When $\beta=70$, both overtakes rate and collisions rate are the best.}}
    \label{fig:beta}
\end{figure}

We train 10 preliminary policies using HG-DAgger with the data filter for different $\beta$ values ranging from 0 to 100 steps and find out that setting $\beta$ as 70 works best for our scenario (both best overtakes rate and collisions rate), which can be found in Figure \ref{fig:beta}. 

\begin{figure}
    \centering
    \includegraphics[width=0.8\columnwidth]{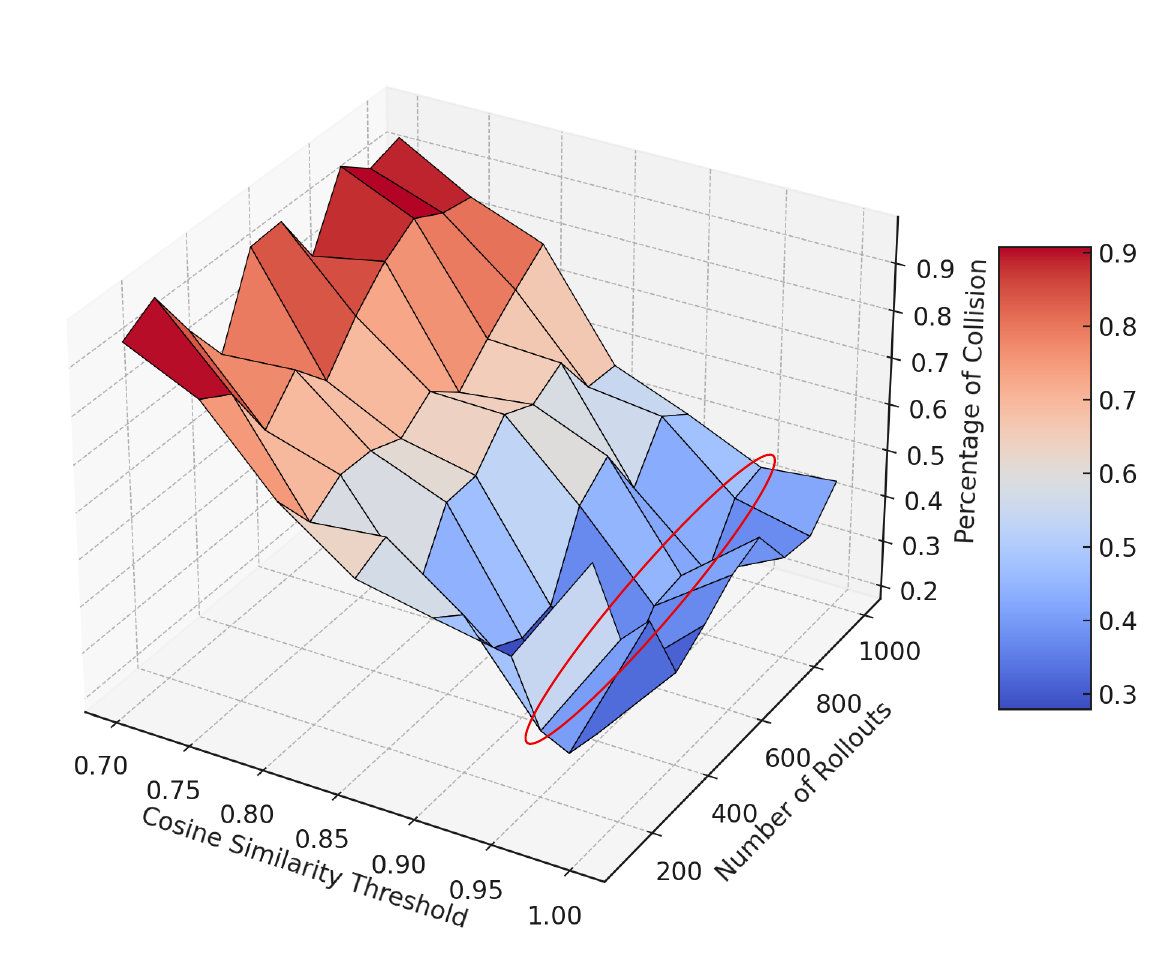}
    \caption{\small{Surface plot that shows the relationship between the cosine similarity threshold, the number of training rollouts, and the percentage of collision. A threshold value of 0.95 works the best in our scenario (marked by the red ellipse).}}
    \label{fig:cosine_sim_collision}
\end{figure}

\begin{figure}
\vspace{-10pt}
    \centering
    \includegraphics[width=0.8\columnwidth]{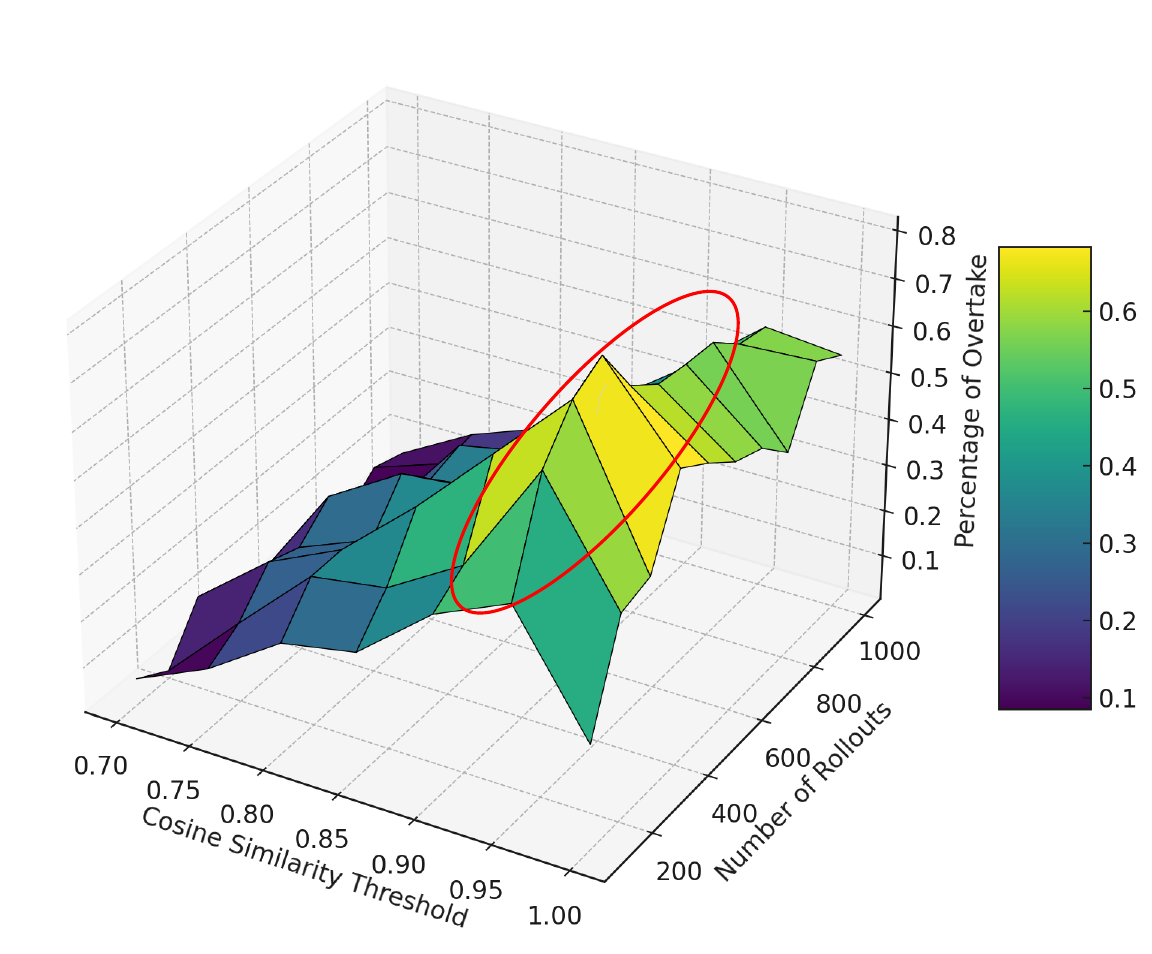}
    \caption{\small{Surface plot displaying the relationship between cosine similarity threshold, number of training rollouts, and percentage of overtake. This plot reveals a trend that complements the result shown in Figure \ref{fig:cosine_sim_collision}.}}
    \label{fig:cosine_sim_overtake}
    \vspace{-20pt}
\end{figure}

% \section*{Acknowledgment}
% The authors thank Derek Zhou and Zhijun Zhuang for their diligent help on the experiments, and Luigi Berducci, Nandan Tumu, and Hongrui Zheng for useful discussion.

\bibliographystyle{unsrt}
\bibliography{references}

\begin{thebibliography}{10}

\bibitem{pierson2017deep}
H.~A. Pierson and M.~S. Gashler.
\newblock Deep learning in robotics: a review of recent research.
\newblock {\em Advanced Robotics}, 31(16):821--835, 2017.

\bibitem{argall2009survey}
B.~D. Argall, S.~Chernova, M.~Veloso, and B.~Browning.
\newblock A survey of robot learning from demonstration.
\newblock {\em Robotics and autonomous systems}, 57(5):469--483, 2009.

\bibitem{fang2019survey}
B.~Fang, S.~Jia, D.~Guo, M.~Xu, S.~Wen, and F.~Sun.
\newblock Survey of imitation learning for robotic manipulation.
\newblock {\em International Journal of Intelligent Robotics and Applications},
  3:362--369, 2019.

\bibitem{le2022survey}
L.~Le~Mero, D.~Yi, M.~Dianati, and A.~Mouzakitis.
\newblock A survey on imitation learning techniques for end-to-end autonomous
  vehicles.
\newblock {\em IEEE Transactions on Intelligent Transportation Systems}, 2022.

\bibitem{bojarski2016end}
M.~Bojarski, D.~Del~Testa, D.~Dworakowski, Be. Firner, B.~Flepp, P.~Goyal,
  L.~D. Jackel, M.~Monfort, U.~Muller, J.~Zhang, et~al.
\newblock End to end learning for self-driving cars.
\newblock {\em arXiv preprint arXiv:1604.07316}, 2016.

\bibitem{ross2011reduction}
S.~Ross, G.~Gordon, and D.~Bagnell.
\newblock A reduction of imitation learning and structured prediction to
  no-regret online learning.
\newblock In {\em Proceedings of the fourteenth international conference on
  artificial intelligence and statistics}, pages 627--635. JMLR Workshop and
  Conference Proceedings, 2011.

\bibitem{kelly2019hg}
M.~Kelly, C.~Sidrane, K.~Driggs-Campbell, and M.~J. Kochenderfer.
\newblock Hg-dagger: Interactive imitation learning with human experts.
\newblock In {\em 2019 International Conference on Robotics and Automation
  (ICRA)}, pages 8077--8083. IEEE, 2019.

\bibitem{ross2010efficient}
S.~Ross and D.~Bagnell.
\newblock Efficient reductions for imitation learning.
\newblock In {\em Proceedings of the thirteenth international conference on
  artificial intelligence and statistics}, pages 661--668. JMLR Workshop and
  Conference Proceedings, 2010.

\bibitem{zhang2016query}
J.~Zhang and K.~Cho.
\newblock Query-efficient imitation learning for end-to-end autonomous driving.
\newblock {\em arXiv preprint arXiv:1605.06450}, 2016.

\bibitem{menda2019ensembledagger}
K.~Menda, K.~Driggs-Campbell, and M.~J. Kochenderfer.
\newblock Ensembledagger: A bayesian approach to safe imitation learning.
\newblock In {\em 2019 IEEE/RSJ International Conference on Intelligent Robots
  and Systems (IROS)}, pages 5041--5048. IEEE, 2019.

\bibitem{hoque2021lazydagger}
R.~Hoque, A.~Balakrishna, C.~Putterman, M.~Luo, D.~S. Brown, D.~Seita,
  B.~Thananjeyan, E.~Novoseller, and K.~Goldberg.
\newblock Lazydagger: Reducing context switching in interactive imitation
  learning.
\newblock In {\em 2021 IEEE 17th International Conference on Automation Science
  and Engineering (CASE)}, pages 502--509. IEEE, 2021.

\bibitem{hoque2021thriftydagger}
R.~Hoque, A.~Balakrishna, E.~Novoseller, A.~Wilcox, D.~S. Brown, and
  K.~Goldberg.
\newblock Thriftydagger: Budget-aware novelty and risk gating for interactive
  imitation learning.
\newblock {\em arXiv preprint arXiv:2109.08273}, 2021.

\bibitem{penndot}
Pennsylvania~Department of~Transportation.
\newblock 2021 pennsylvania crash facts \& statistics.
\newblock
  \url{https://www.penndot.pa.gov/TravelInPA/Safety/Documents/2021_CFB_linked.pdf}.

\bibitem{abbeel2004apprenticeship}
P.~Abbeel and A.~Y. Ng.
\newblock Apprenticeship learning via inverse reinforcement learning.
\newblock In {\em Proceedings of the twenty-first international conference on
  Machine learning}, page~1, 2004.

\bibitem{pan2017agile}
Y.~Pan, C.-A. Cheng, K.~Saigol, K.~Lee, X.~Yan, E.~Theodorou, and B.~Boots.
\newblock Agile autonomous driving using end-to-end deep imitation learning.
\newblock {\em arXiv preprint arXiv:1709.07174}, 2017.

\bibitem{zhou2021exploring}
J.~Zhou, R.~Wang, X.~Liu, Y.~Jiang, S.~Jiang, J.~Tao, J.~Miao, and S.~Song.
\newblock Exploring imitation learning for autonomous driving with feedback
  synthesizer and differentiable rasterization.
\newblock In {\em 2021 IEEE/RSJ International Conference on Intelligent Robots
  and Systems (IROS)}, pages 1450--1457. IEEE, 2021.

\bibitem{sun2022benchmark}
Xiatao Sun, Mingyan Zhou, Zhijun Zhuang, Shuo Yang, Johannes Betz, and Rahul
  Mangharam.
\newblock A benchmark comparison of imitation learning-based control policies
  for autonomous racing.
\newblock In {\em 2023 IEEE Intelligent Vehicles Symposium (IV)}, pages 1--5,
  2023.

\bibitem{wu2019imitation}
Y.-H. Wu, N.~Charoenphakdee, H.~Bao, V.~Tangkaratt, and M.~Sugiyama.
\newblock Imitation learning from imperfect demonstration.
\newblock In {\em International Conference on Machine Learning}, pages
  6818--6827. PMLR, 2019.

\bibitem{brown2020better}
D.~S. Brown, W.~Goo, and S.~Niekum.
\newblock Better-than-demonstrator imitation learning via automatically-ranked
  demonstrations.
\newblock In {\em Conference on robot learning}, pages 330--359. PMLR, 2020.

\bibitem{wang2021learning}
Y.~Wang, C.~Xu, B.~Du, and H.~Lee.
\newblock Learning to weight imperfect demonstrations.
\newblock In {\em International Conference on Machine Learning}, pages
  10961--10970. PMLR, 2021.

\bibitem{brown2019extrapolating}
D.~Brown, W.~Goo, P.~Nagarajan, and S.~Niekum.
\newblock Extrapolating beyond suboptimal demonstrations via inverse
  reinforcement learning from observations.
\newblock In {\em International conference on machine learning}, pages
  783--792. PMLR, 2019.

\bibitem{raykar2009supervised}
V.~C. Raykar, S.~Yu, L.~H. Zhao, A.~Jerebko, C.~Florin, G.~H. Valadez,
  L.~Bogoni, and L.~Moy.
\newblock Supervised learning from multiple experts: whom to trust when
  everyone lies a bit.
\newblock In {\em Proceedings of the 26th Annual international conference on
  machine learning}, pages 889--896, 2009.

\bibitem{raykar2010learning}
V.~C. Raykar, S.~Yu, L.~H. Zhao, G.~H. Valadez, C.~Florin, L.~Bogoni, and
  L.~Moy.
\newblock Learning from crowds.
\newblock {\em Journal of machine learning research}, 11(4), 2010.

\bibitem{betz2022autonomous}
J.~Betz, H.~Zheng, A.~Liniger, U.~Rosolia, P.~Karle, M.~Behl, V.~Krovi, and
  R.~Mangharam.
\newblock Autonomous vehicles on the edge: A survey on autonomous vehicle
  racing.
\newblock {\em IEEE Open Journal of Intelligent Transportation Systems},
  3:458--488, 2022.

\bibitem{wischnewski2022indy}
A.~Wischnewski, M.~Geisslinger, J.~Betz, T.~Betz, F.~Fent, A.~Heilmeier,
  L.~Hermansdorfer, T.~Herrmann, S.~Huch, P.~Karle, et~al.
\newblock Indy autonomous challenge-autonomous race cars at the handling
  limits.
\newblock In {\em 12th International Munich Chassis Symposium 2021: chassis.
  tech plus}, pages 163--182. Springer, 2022.

\bibitem{rieber2004roborace}
J.~M. Rieber, H.~Wehlan, and F.~Allgower.
\newblock The roborace contest.
\newblock {\em IEEE Control Systems Magazine}, 24(5):57--60, 2004.

\bibitem{zeilinger2017design}
M.~Zeilinger, R.~Hauk, M.~Bader, and A.~Hofmann.
\newblock Design of an autonomous race car for the formula student driverless
  (fsd).
\newblock In {\em Oagm \& Arw Joint Workshop}, 2017.

\bibitem{o2020f1tenth}
M.~O'Kelly, H.~Zheng, D.~Karthik, and R.~Mangharam.
\newblock F1tenth: An open-source evaluation environment for continuous control
  and reinforcement learning.
\newblock {\em Proceedings of Machine Learning Research}, 123, 2020.

\bibitem{herman2021learn}
J.~Herman, J.~Francis, S.~Ganju, B.~Chen, A.~Koul, A.~Gupta, A.~Skabelkin,
  I.~Zhukov, M.~Kumskoy, and E.~Nyberg.
\newblock Learn-to-race: A multimodal control environment for autonomous
  racing.
\newblock In {\em Proceedings of the IEEE/CVF International Conference on
  Computer Vision}, pages 9793--9802, 2021.

\bibitem{kabzan2019learning}
J.~Kabzan, L.~Hewing, A.~Liniger, and M.~N. Zeilinger.
\newblock Learning-based model predictive control for autonomous racing.
\newblock {\em IEEE Robotics and Automation Letters}, 4(4):3363--3370, 2019.

\bibitem{ames2016control}
A.~D. Ames, X.~Xu, J.~W. Grizzle, and P.~Tabuada.
\newblock Control barrier function based quadratic programs for safety critical
  systems.
\newblock {\em IEEE Transactions on Automatic Control}, 62(8):3861--3876, 2016.

\bibitem{zeng2021enhancing}
J.~Zeng, Z.~Li, and K.~Sreenath.
\newblock Enhancing feasibility and safety of nonlinear model predictive
  control with discrete-time control barrier functions.
\newblock In {\em 2021 60th IEEE Conference on Decision and Control (CDC)},
  pages 6137--6144. IEEE, 2021.

\bibitem{liu2019pedestrian}
K.~Liu, W.~Wang, and J.~Wang.
\newblock Pedestrian detection with lidar point clouds based on single template
  matching.
\newblock {\em Electronics}, 8(7):780, 2019.

\bibitem{yang2023global}
C.~Yang, Z.~Zhou, H.~Zhuang, C.~Wang, and M.~Yang.
\newblock Global pose initialization based on gridded gaussian distribution
  with wasserstein distance.
\newblock {\em IEEE Transactions on Intelligent Transportation Systems}, pages
  1--11, 2023.

\bibitem{zhou2015sharp}
J.~Zhou, J.~Yan, T.~Wei, K.~Wu, X.~Chen, and S.~Hu.
\newblock Sharp corner/edge recognition in domestic environments using rgb-d
  camera systems.
\newblock {\em IEEE Transactions on Circuits and Systems II: Express Briefs},
  62(10):987--991, 2015.

\bibitem{hu2022we}
Z.~Hu, D.~Yang, S.~Cheng, L.~Zhou, S.~Wu, and J.~Liu.
\newblock We know where they are looking at from the rgb-d camera: Gaze
  following in 3d.
\newblock {\em IEEE Transactions on Instrumentation and Measurement}, 71:1--14,
  2022.

\bibitem{f1tenthicra2022results}
{F1TENTH ICRA 2022}: Results.
\newblock \url{https: //icra2022-race.f1tenth.org/results.html}.

\end{thebibliography}

\end{document}